\documentclass[review]{elsarticle}
\usepackage[numbers]{natbib}
\usepackage{graphicx}
\usepackage{float}
\usepackage{booktabs}
\usepackage{multirow}
\usepackage{makecell}
\usepackage{subfigure}
\usepackage{array}
\usepackage{amsmath,amssymb}
\usepackage{xcolor}
\usepackage{pifont}
\usepackage{bm}
\usepackage{enumerate}
\usepackage{siunitx}
\usepackage{lmodern}
\usepackage{textcomp}
\usepackage{adjustbox}
\usepackage{wasysym}
\usepackage{hyperref}
\graphicspath{ {./Picture/} }
\restylefloat{figure}
\floatstyle{plaintop}
\restylefloat{table}

\journal{Robotics and Computer-Integrated Manufacturing}

\begin{document}
\begin{frontmatter}

\title{Multi-Period Texture Contrast Enhancement for Low-Contrast Wafer Defect Detection and Segmentation}

\author[aff1]{Zihan Zhang}

\author[aff1]{Shijiao Li\corref{cor1}}
\ead{shi-jiao@outlook.com}

\author[aff1]{Danhang Niu}

\author[aff1]{Wei Peng\corref{cor2}}
\ead{pw2022@zzu.edu.cn}

\author[aff2]{Yifan Hu\corref{cor3}}
\ead{yifan-hu019@163.com}

\author[aff1,aff3]{Mingfu Zhu\corref{cor4}}
\ead{zhumingfu@zzu.edu.cn}

\author[aff4]{Rui Xi}

\author[aff5]{Lixin Zhang}

\author[aff6]{Tien-Chien Jen\corref{cor5}}
\ead{tjen@uj.ac.za}

\author[aff1,aff6]{Ronghan Wei\corref{cor6}}
\ead{prof.wei@outlook.com}

\cortext[cor1,cor2,cor3,cor4,cor5,cor6]{Corresponding authors.}

\address[aff1]{Engineering Technology Research Center of Henan Province for MEMS Manufacturing and Applications, School of Mechanics and Safety Engineering, Zhengzhou University, Zhengzhou 450001, China}

\address[aff2]{Zhengzhou V3 Biotechnology Co., Ltd., Zhengzhou 450047, China}

\address[aff3]{Institute of Intelligent Sensing, Zhengzhou University, Zhengzhou 450001, China}

\address[aff4]{School of Mechanical Engineering, North China University of Water Resources and Electric Power, Zhengzhou 450001, China}

\address[aff5]{Institute of Bingtuan Energy Development Research, School of Energy and Materials, Shihezi University, Shihezi 83200, China}

\address[aff6]{Mechanical Engineering Science, University of Johannesburg, Cnr Kingsway and University Road, Auckland Park, Johannesburg 2092, South Africa}

\begin{abstract}
Wafer defect segmentation is pivotal for semiconductor yield optimization yet remains challenged by the intrinsic conflict between microscale anomalies and highly periodic, overwhelming background textures. Existing deep learning paradigms often falter due to feature dilution during downsampling and the lack of explicit mechanisms to disentangle low-contrast defects from process-induced noise. To transcend these limitations, we propose \textbf{TexWDS}, a texture-aware framework that harmonizes multi-scale feature retention with frequency-domain perturbation modeling. Our methodology incorporates three strategic innovations: (1) A Multi-scale Receptive Field Reweighting strategy is introduced to mitigate aliasing effects and preserve high-frequency details of micro-defects often lost in standard pyramidal architectures. (2) The Multi-scale Unified Semantic Enhancer (MUSE) integrates local appearance with global context encoding, effectively enhancing feature discriminability in low-visibility regions. (3) Crucially, we design a plug-and-play Multi-Periodic Texture Contrast Enhancement (MPTCE) module. By modeling texture disruptions in the frequency domain, MPTCE explicitly decouples non-periodic anomalies from structured backgrounds, boosting contrast for camouflaged defects. Extensive experiments on real-world industrial datasets demonstrate that TexWDS achieves a new state-of-the-art, surpassing the baseline by 8.3\% in $mAP_{50-95}$ and 7.7\% in recall, while reducing the false positive rate by approximately 8.6\%. These results underscore the framework's robustness in handling complex periodic patterns and its suitability for high-precision manufacturing inspection.
\end{abstract}


\end{frontmatter}

\section{Introduction}
\label{sec:introduction}

In the realm of semiconductor manufacturing, the precise pixel-level detection and segmentation of wafer defects are paramount for yield management and process control~\cite{Kim2023}. As semiconductor device nodes continue to shrink, defects such as micro-scratches, contamination, and etching anomalies have become increasingly subtle, rendering conventional inspection methods~\cite{haralick1973textural,10.1109/TPAMI.1986.4767851} inadequate. Consequently, there is a heightened reliance on automated defect inspection systems driven by advanced computer vision techniques~\cite{LI2023102470RCIM, WANG2023102513RCIM, WANG2024102791RCIM}.

Recent years have witnessed remarkable progress in deep learning across various vision tasks, offering valuable insights for high-precision industrial inspection. For instance, advanced generative models have achieved breakthroughs in fine-grained texture generation~\cite{shen2025imaggarment}, controllable layout consistency~\cite{shen2025imagharmony}, and complex subject transformation~\cite{shen2025imagedit}. Furthermore, the utilization of structural priors has proven effective in modeling long-term dependencies in dynamic scenes~\cite{shenlong}. Despite these general advancements, applying standard deep learning frameworks, such as Convolutional Neural Networks (CNNs)~\cite{8263132,mi14050905, article, INGLE2025103158} and the YOLO series~\cite{7780444,10295689, Terven_2023,10.1117/12.3033091}, to the specific domain of wafer inspection remains fraught with challenges.

\begin{figure}[t]
    \centering
    \includegraphics[width=1\linewidth]{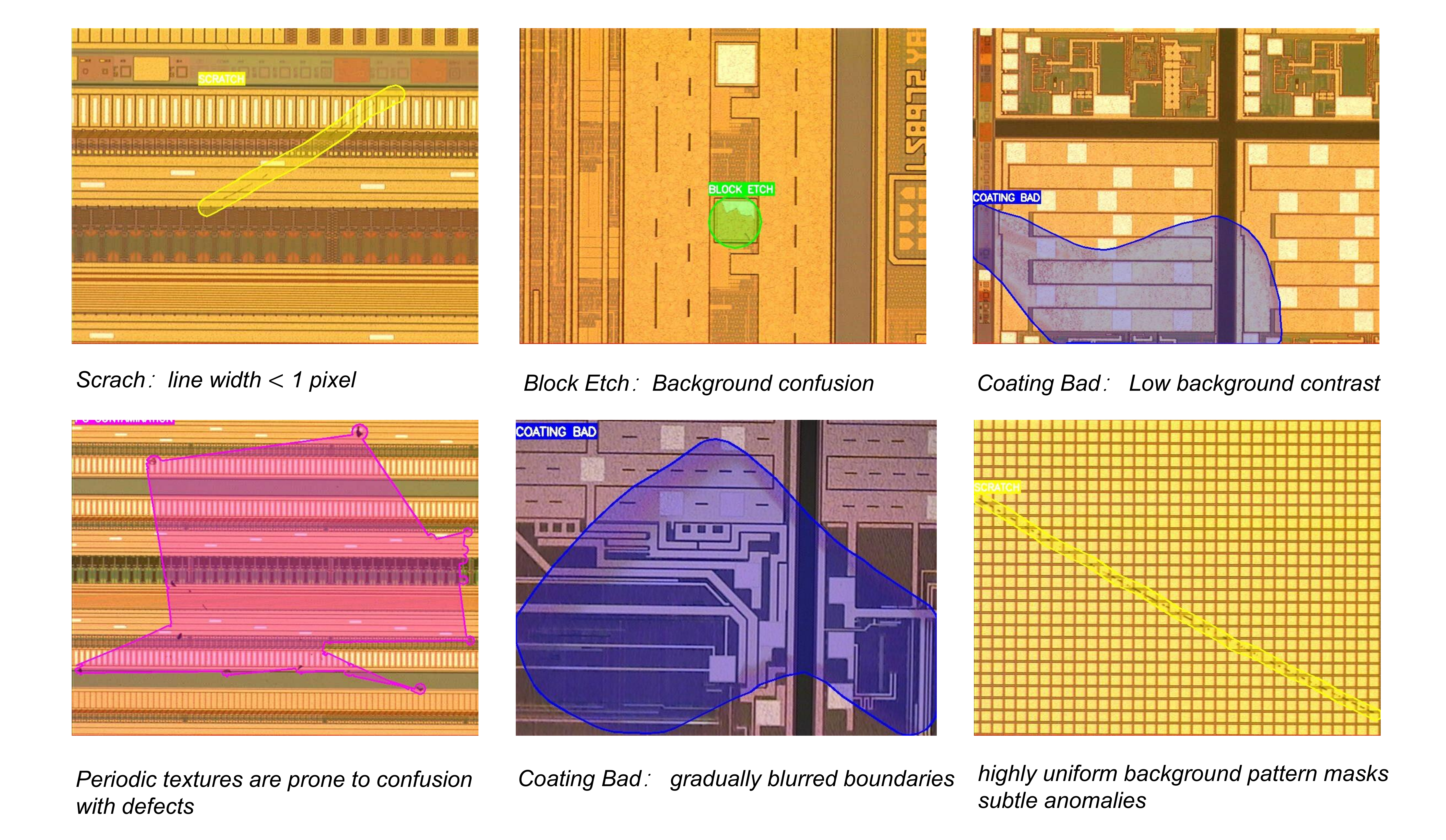}
    \caption{Examples of scenarios where existing deep learning models perform suboptimally.}
    \label{fig:examples}
\end{figure}
      
Current mainstream models often falter when transposed to wafer imagery due to three fundamental limitations. First, existing architectures often underestimate the importance of fine-grained patterns, leading to suboptimal recall for subpixel anomalies near the resolution limit~\cite{8099490,8578843,8746000}. Second, large-area defects with diffuse boundaries are often confused with process noise, leading to high false-negative rates. Third, the highly periodic, repetitive circuit textures that dominate wafer surfaces serve as sophisticated camouflage. Standard models often lack the specific texture-awareness required to disentangle these periodic patterns from true anomalies. These limitations highlight a significant gap in current methodologies and underscore the urgent need for a robust framework that is explicitly context-aware and texture-sensitive~\cite{7515504,9090668}, capable of accounting for the unique statistical and structural properties of wafer imagery.

\begin{figure}[t]
    \centering
    \includegraphics[width=1\linewidth]{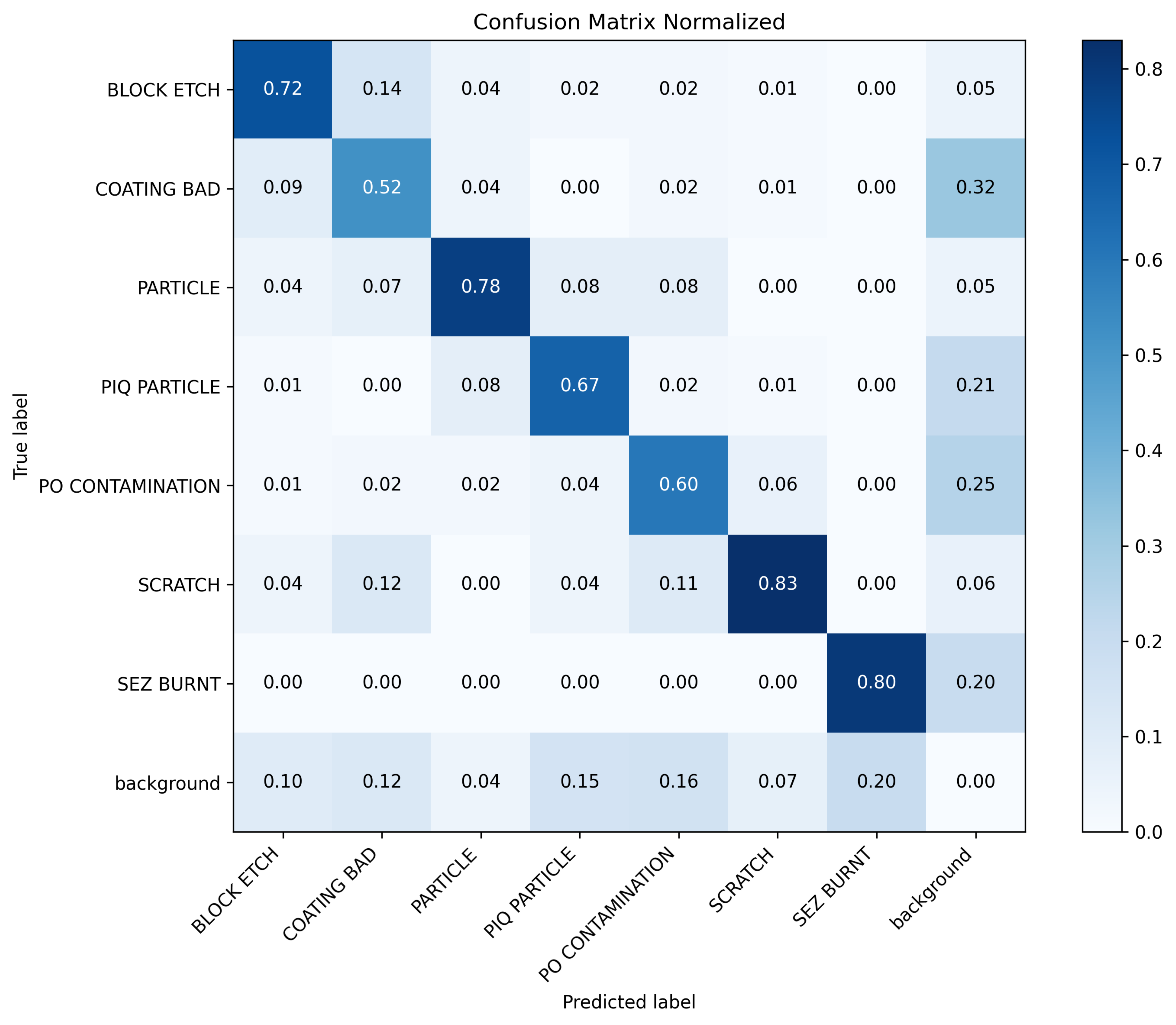}
    \caption{Normalized confusion matrix of YOLOv8-seg (baseline).}
    \label{fig:confusion_matrix}
\end{figure}

\begin{figure}[t]
    \centering
    \includegraphics[width=1\linewidth]{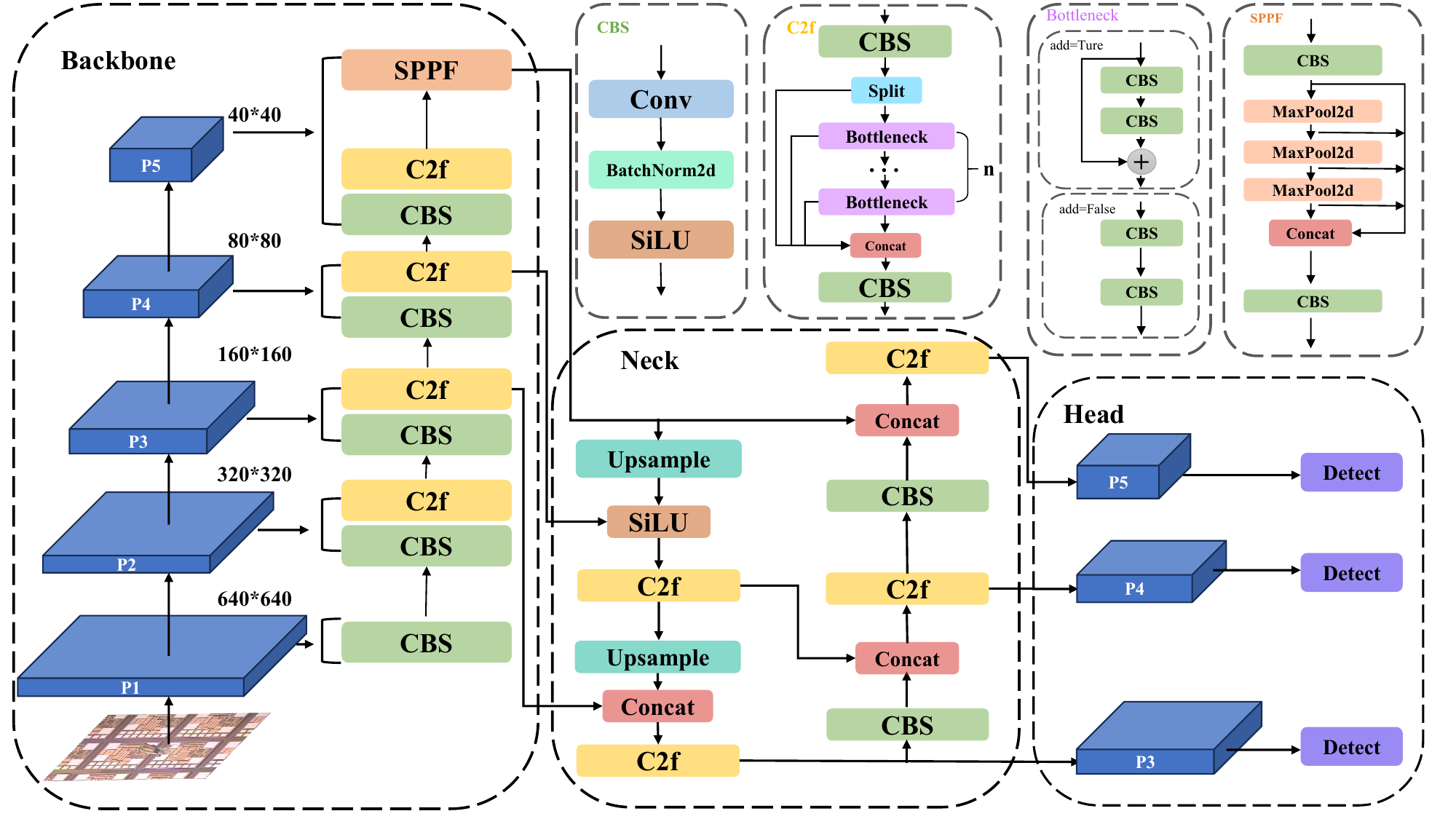}
    \caption{Schematic overview of the YOLOv8-seg (baseline) architecture.}
    \label{fig:baseline}
\end{figure}

To surmount these obstacles, we propose Texture-Aware Wafer Defect Segmentation (TexWDS), an enhanced framework built upon YOLOv8-seg. TexWDS incorporates architectural and representational innovations designed to bridge the gap between general detection models and the intricate physical properties of semiconductor wafers. Our approach focuses on enhancing feature discriminability under low-visibility conditions through three strategic contributions.
We first introduce a Multi-scale Receptive Field Reweighting strategy~\cite{chen2018encoder}. This module adaptively emphasizes defect-rich regions while suppressing interference from repetitive background patterns. By recalibrating feature responses, it significantly enhances the model's sensitivity to small-scale, low-contrast anomalies, effectively mitigating feature underrepresentation in low-visibility regions.
Additionally, we propose the Multi-scale Unified Semantic Enhancer with Context Encoding (MUSE). Relying solely on local pixel information often leads to ambiguity and false positives. To address this, MUSE fuses local appearance features with global contextual semantics~\cite{Long_2015_CVPR, Lin_2017_CVPR,8237467}. This integration enables the model to distinguish true defects from process-induced noise by interpreting the structural abnormality of a candidate region within the full wafer context.
Finally, we integrate a plug-and-play Multi-Periodic Texture Contrast Enhancement (MPTCE) module. Drawing inspiration from frequency domain analysis, this module is explicitly designed to model and amplify periodic texture deviations across multiple scales. By isolating non-periodic anomalies from structured background textures, MPTCE significantly boosts the detection of camouflaged defects that are frequently missed by conventional approaches.

We rigorously evaluate our method on a proprietary wafer defect dataset encompassing seven defect categories commonly encountered in semiconductor manufacturing. Extensive experiments demonstrate that TexWDS significantly outperforms state-of-the-art methods. Notably, integrating MPTCE yields substantial improvements in both recall and precision, confirming the effectiveness of our approach in handling complex wafer surface textures.

The main contributions of this paper are summarized as follows:
\begin{itemize}
    \item We propose a novel framework, TexWDS, that integrates multi-scale receptive field reweighting and local-global context fusion to enhance defect sensitivity while efficiently suppressing periodic background interference.
    \item We introduce a plug-and-play MPTCE module that explicitly models microscale periodic disturbances in the frequency domain, significantly improving discriminability for low-contrast defects without the need for extensive retraining.
    \item Extensive experiments on industrial wafer defect datasets demonstrate state-of-the-art performance, achieving +8.3 percentage points in mAP$_{50-95}$ and +7.7 percentage points in recall, surpassing existing methods in challenging low-contrast and large-area defect scenarios.
\end{itemize}

\section{Related Work}

\subsection{Conventional Wafer Inspection Algorithms}
Early wafer inspection primarily relied on template matching and morphological processing. Standard techniques involve referencing golden templates to identify discrepancies via subtraction or statistical thresholding. While edge detection operators (e.g., Canny~\citep{10.1109/TPAMI.1986.4767851}, Sobel) effectively delineate prominent boundaries, they rely heavily on heuristic thresholds, making them brittle to process variations. To address surface textures, statistical methods employing Gabor filters and Gray-Level Co-occurrence Matrices (GLCM)~\citep{haralick1973textural} were introduced to model local spatial dependencies. However, these handcrafted feature descriptors lack semantic adaptability, struggling to differentiate defects from complex periodic circuit patterns or to detect microscale anomalies with weak intensity contrast~\citep{Ma2023Review}.

\subsection{Deep Learning-Based Generic Detection}
The advent of Convolutional Neural Networks (CNNs) has revolutionized defect inspection~\citep{Dey2022MaskRCNNSEM, LopezDeLaRosa2023SqueezeNet}. Object detectors like the YOLO series~\citep{redmon2018yolov3incrementalimprovement,bochkovskiy2020yolov4optimalspeedaccuracy} and segmentation networks like DeepLabv3~\citep{chen2017rethinkingatrousconvolutionsemantic} have been widely adapted for industrial tasks due to their robust feature extraction capabilities~\citep{Tyrone2023Benchmarking}. Recent advances in general vision, such as complex subject transformation~\citep{shen2025imagedit}, further demonstrate the potential of deep representations in capturing diverse visual morphologies. To handle scale variations, Feature Pyramid Networks (FPN)~\citep{lin2017featurepyramidnetworksobject} are commonly employed to aggregate multi-level semantics. Despite their success, these generic models often falter in wafer inspection: diverse pooling operations can dilute feature responses to sub-pixel defects, and standard convolutions lack mechanisms to suppress overwhelming interference from high-frequency repetitive backgrounds, leading to high false-negative rates for low-contrast targets.

\subsection{Context-Aware and Texture-Specific Modeling}
Recognizing the limitations of generic models, recent research focuses on enhancing feature discriminability through attention mechanisms and structural priors. Analogous to how motion priors model long-term dependencies in dynamic scenes~\citep{shen2025imaggarment}, spatial context modeling (e.g., CGNet~\citep{9292449}, GCNet~\citep{cao2020globalcontextnetworks}) refines local representations with global semantic constraints. Attention modules, such as RCAN~\citep{zhang2018imagesuperresolutionusingdeep} and CBAM, have been integrated into inspection networks (e.g., YOLOv8-ResCBAM~\citep{YAN2024e30889}) to recalibrate feature responses, emphasizing informative regions while suppressing noise. Furthermore, breakthroughs in fine-grained texture generation~\citep{shen2024imagpose,shen2024advancing} and controllable layout consistency~\citep{shen2025imagdressing,shenlong} underscore the need for explicit modelling of texture and structure to handle complex visual patterns. Inspired by these developments, yet tailored to the specific domain of wafer inspection, our \textbf{TexWDS} framework introduces a frequency-domain enhancement module. Unlike previous approaches, it explicitly decouples defects from periodic backgrounds, ensuring robust segmentation even under extremely low-contrast conditions.

\section{Method}
\begin{figure}[t]
    \centering
    \includegraphics[width=1\linewidth]{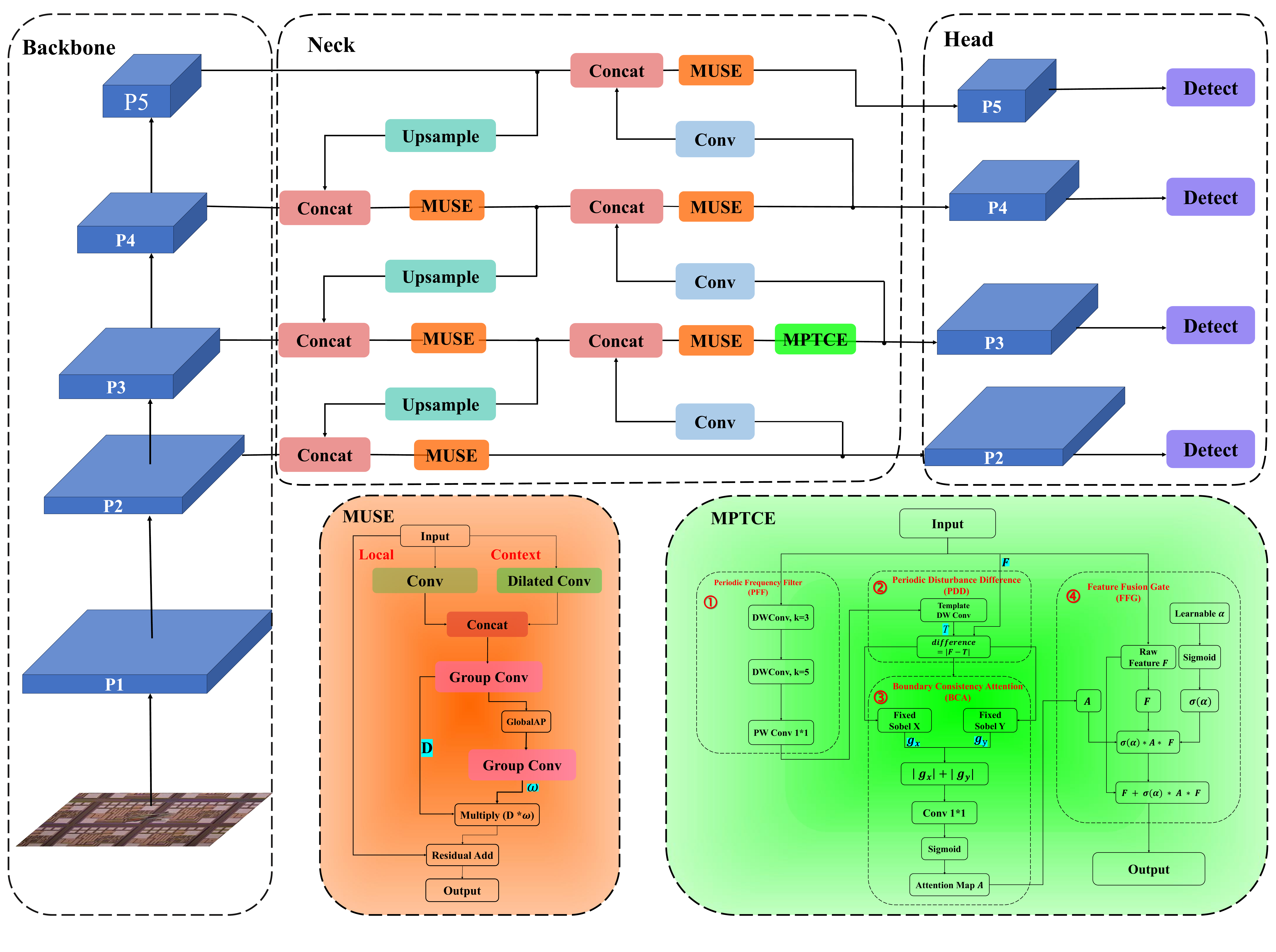}
    \caption{Schematic overview of the proposed Texture-Aware Wafer Defect Segmentation (TexWDS) architecture, showing the integration positions of Stage~1 (P2 branch), Stage~2 (Multi-scale Unified Semantic Enhancer with Context Encoding (MUSE), and Stage~3 (Multi-Period Texture Contrast Enhancement, MPTCE) within the backbone--neck pipeline.}
    \label{fig: overall}
\end{figure}
The overall architecture of our module TexWDS is illustrated in Fig. ~\ref{fig: overall} and Table ~\ref{tab: para of TexWDS}. The proposed framework achieves collaborative modeling of spatial scale, contextual consistency, and periodic texture perturbations. Key design principles are summarized as follows:

\noindent\textbf{Stage 1 Multi-scale Receptive Field Reweighting:} Enhance geometric details by emphasizing the spatial structures of micro-particles and fine scratches.

\noindent\textbf{Stage 2 Multi-scale Unified Semantic Enhancer with Context Encoding (MUSE):} Strengthen contextual discriminability to resolve confusion between low-contrast backgrounds and contamination-type defects.

\noindent\textbf{Stage 3 MPTCE:} Explicitly model periodic texture perturbations to address the challenge where defects closely resemble periodic circuit patterns.

The three modules jointly form a tri-domain enhancement structure: the geometric detail domain, the contextual consistency domain, and the periodic disturbance domain.
The backbone features are decomposed into three disentangled components:
\begin{equation}
    \mathbf{f}_{\text{geom}}, \quad \mathbf{f}_{\text{context}}, \quad \mathbf{f}_{\text{texture}} .
\end{equation}
Each of which is selectively enhanced by its corresponding module and fused into a unified representation:
\begin{equation}
    \mathbf{f}_{\text{final}} = g\!\left(\mathbf{f}_{\text{geom}}, \mathbf{f}_{\text{context}}, \mathbf{f}_{\text{texture}}\right) .
\end{equation}
where $g(\cdot)$ denotes the feature fusion function.

\begin{table}[htbp]
\centering
\caption{The parameters of the proposed TexWDS model.}
\label{tab: para of TexWDS}
\footnotesize 
\renewcommand{\arraystretch}{0.9} 
\resizebox{\linewidth}{!}{
\begin{tabular}{l c c c c c}
\toprule
\textbf{Layer} & \textbf{Input Size} & \textbf{Filter Size} & \textbf{Filter Num} & \textbf{Stride} & \textbf{Padding} \\
\midrule
Conv & $640\times640$ & $3\times3$ & 64 & 2 & 1 \\
Conv & $320\times320$ & $3\times3$ & 128 & 2 & 1 \\
C2f\_MUSE & $160\times160$ &
$\left[\begin{array}{c}
1\times1,64 \\
3\times3,64
\end{array}\right]\times3$ & 128 & 1 & 1 \\
Conv & $160\times160$ & $3\times3$ & 256 & 2 & 1 \\
C2f\_MUSE & $80\times80$ &
$\left[\begin{array}{c}
1\times1,128 \\
3\times3,128
\end{array}\right]\times6$ & 256 & 1 & 1 \\
Conv & $80\times80$ & $3\times3$ & 512 & 2 & 1 \\
C2f\_MUSE & $40\times40$ &
$\left[\begin{array}{c}
1\times1,256 \\
3\times3,256
\end{array}\right]\times6$ & 512 & 1 & 1 \\
Conv & $40\times40$ & $3\times3$ & 1024 & 2 & 1 \\
C2f & $20\times20$ &
$\left[\begin{array}{c}
1\times1,512 \\
3\times3,512
\end{array}\right]\times2$ & 1024 & 1 & 1 \\
SPPF & $20\times20$ & MaxPool $5\times5$ & 1024 & 1 & 2 \\
Upsample & $20\to40$ & — & 1024 & — & — \\
Concat & $40\times40$ & — & 1536 & — & — \\
C2f\_MUSE & $40\times40$ &
$\left[\begin{array}{c}
1\times1,256 \\
3\times3,256
\end{array}\right]\times3$ & 512 & 1 & 1 \\
Upsample & $40\to80$ & — & 512 & — & — \\
Concat & $80\times80$ & — & 768 & — & — \\
C2f\_MUSE & $80\times80$ &
$\left[\begin{array}{c}
1\times1,128 \\
3\times3,128
\end{array}\right]\times3$ & 256 & 1 & 1 \\
Upsample & $80\to160$ & — & 256 & — & — \\
Concat & $160\times160$ & — & 384 & — & — \\
C2f\_MUSE & $160\times160$ &
$\left[\begin{array}{c}
1\times1,64 \\
3\times3,64
\end{array}\right]\times3$ & 128 & 1 & 1 \\
Conv & $160\times160$ & $3\times3$ & 128 & 2 & 1 \\
Concat & $80\times80$ & — & 384 & — & — \\
C2f\_MUSE & $80\times80$ &
$\left[\begin{array}{c}
1\times1,128 \\
3\times3,128
\end{array}\right]\times3$ & 256 & 1 & 1 \\
MPTCE & $80\times80$ & Multi-scale Pyramid TCN & 256 & 1 & — \\
Conv & $80\times80$ & $3\times3$ & 256 & 2 & 1 \\
Concat & $40\times40$ & — & 768 & — & — \\
MUSE & $40\times40$ &
$\left[\begin{array}{c}
1\times1,256 \\
3\times3,256 \\
\text{Channel Attention}
\end{array}\right]\times2$ & 512 & 1 & 1 \\
Conv & $40\times40$ & $3\times3$ & 512 & 2 & 1 \\
Concat & $20\times20$ & — & 1536 & — & — \\
C2f\_MUSE & $20\times20$ &
$\left[\begin{array}{c}
1\times1,512 \\
3\times3,512
\end{array}\right]\times3$ & 1024 & 1 & 1 \\
Conv (1×1) & $20\times20$ & $1\times1$ & 1024 & 1 & 0 \\
DWConv & $20\times20$ & $3\times3$ (depthwise) & 1024 & 1 & 1 \\
Upsample & $20\to40$ & — & 1024 & — & — \\
Conv (3×3) & $40\times40$ & $3\times3$ & 1024 & 1 & 1 \\
EffectiveSE & $20\times20$ & Channel Attention & 1024 & 1 & — \\
Segment Head & $[160,80,40,20]$ & — & $nc=7$, proto=256 & — & — \\
\bottomrule
\end{tabular}
}
\end{table}
\subsection{Multi-scale Receptive Field Reweighting}
\begin{figure}[t]
    \centering
    \includegraphics[width=1.2\linewidth]{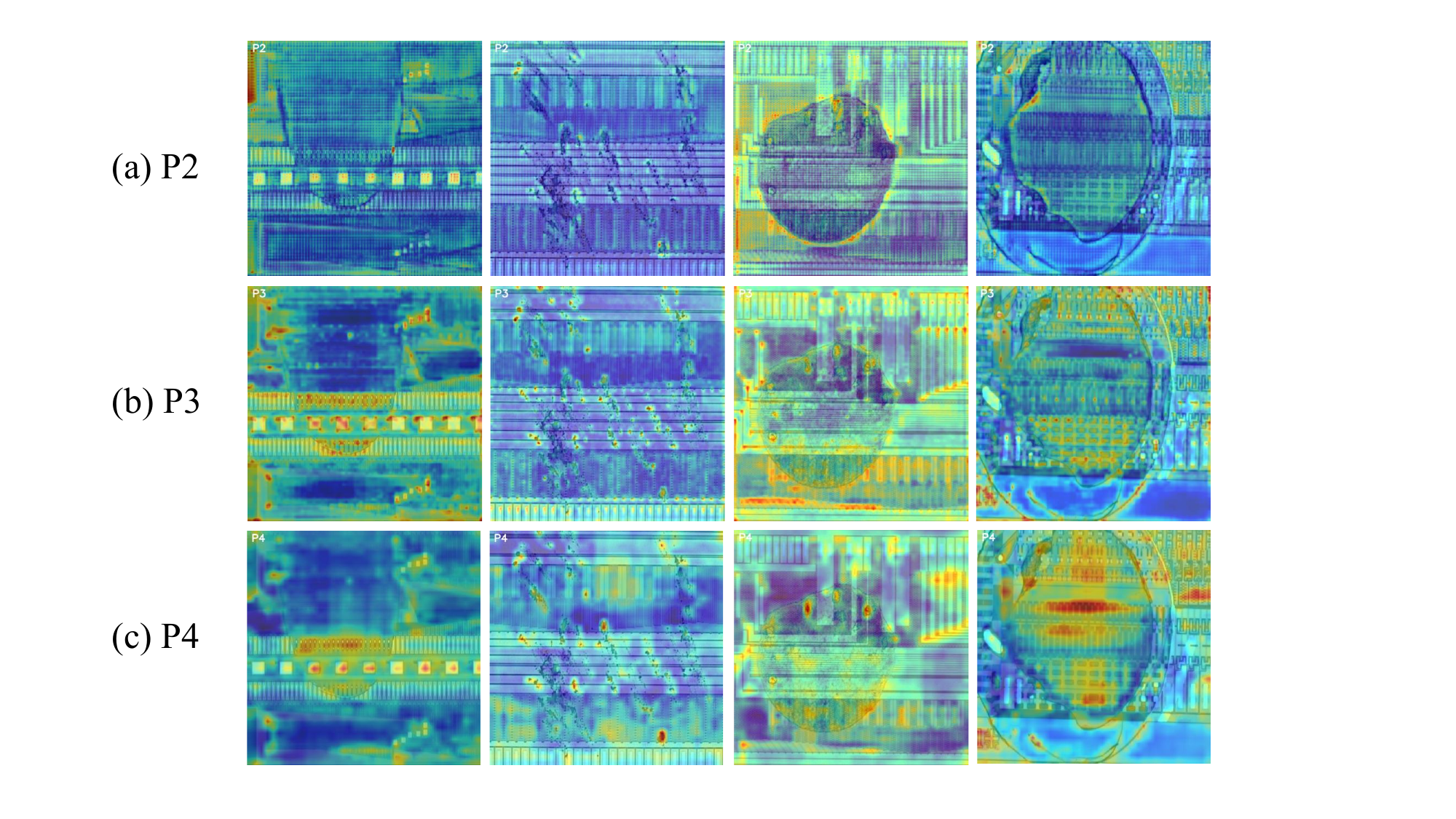}
    \caption{Comparison of feature maps from detection heads at different scales. 
(a), (b) and (c) represent the feature maps of the YOLOv8 model at the P2, P3, and P4 scales, respectively.}
    \label{fig: p2p3p4}
\end{figure}
Common wafer defects, such as linear scratches and localized burnouts, rely heavily on high-frequency edges and micro-structural textures for accurate detection. However, in the original YOLOv8 architecture, these fine-grained features are progressively smoothed or fragmented during the downsampling stages at P3 ($\times 8$) and P4 ($\times 16$) due to insufficient spatial resolution.
Let $w$ denote the effective spatial width (in pixels) of a defect pattern in the input image. After $\times 8$ downsampling, the corresponding width at the P3 level becomes
\begin{equation}
    \frac{w}{8} < 1 .
\end{equation}
where a value below one pixel violates the Nyquist sampling criterion, inevitably leading to aliasing and texture collapse. This observation motivates the introduction of a higher-resolution feature branch dedicated to preserving high-frequency geometric details.

To address this issue, a P2 feature branch with $4\times$ downsampling is introduced at the front end of the backbone network. Fig. ~\ref{fig: p2p3p4} shows the feature maps captured by detection heads at different scales. It can be observed that detection heads with smaller scales are better at identifying defect edges and fine background textures. This branch consists of:
a $1\times 1$ pointwise convolution for channel alignment,
an upsampling operation to match spatial resolution,
a fusion operation with the P3 feature map,
subsequent integration into the path aggregation network (Neck).
Let $\mathbf{C}_2 \in \mathbb{R}^{H\times W\times C}$ denote the backbone feature map at stage~2, and let $\mathbf{P}_3$ denote the corresponding P3 feature map. The operations are defined as:
\begin{equation}
\begin{aligned}
    \mathbf{P}_2 &= \mathrm{Conv}_{1\times 1}(\mathbf{C}_2), \\
    \tilde{\mathbf{P}}_2 &= \mathrm{Upsample}(\mathbf{P}_2), \\
    \mathbf{F}_{\mathrm{P2\text{-}P3}} &= \tilde{\mathbf{P}}_2 \oplus \mathbf{P}_3 .
\end{aligned}
\end{equation}
where $\oplus$ denotes element-wise addition and $\mathbf{F}_{\mathrm{P2\text{-}P3}}$ represents the fused feature map. This design enables the network to preserve continuous linear structures, subtle texture gradients, and sharp edge phase information.

\subsection{Multi-scale Unified Semantic Enhancer with Context Encoding (MUSE)}
\begin{figure*}
    \centering
    \includegraphics[width=1\linewidth]{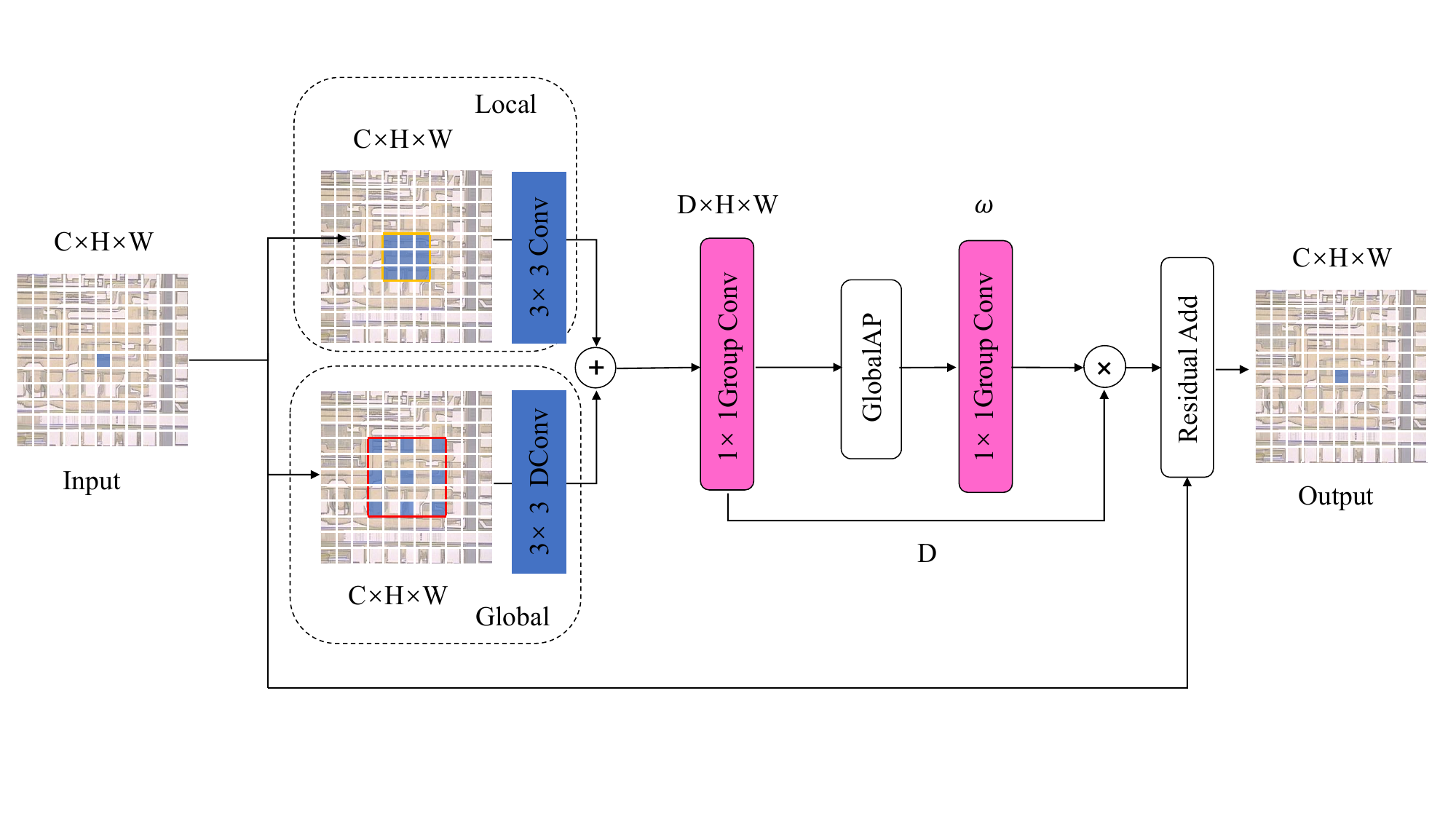}
    \caption{Illustration of the context structure within the proposed MUSE module, which contains three sub-branches: local, surrounding, and global channel attention.}
    \label{fig: module_B}
\end{figure*}

Contamination-type defects and PI particle defects typically exhibit low contrast, blurred boundaries, partial occlusion by periodic wiring backgrounds, and weak local responses. Without sufficient contextual cues, these defects are easily misclassified as background textures. To alleviate this issue, contextual information must be explicitly incorporated to enhance semantic discriminability. The overall structure of the MUSE module is illustrated in Fig.~\ref{fig: module_B}.

Our MUSE module is inspired by the Context Guided (CG) block proposed in CGNet \citep{9292449}.
Different from CGNet, which targets lightweight semantic segmentation in natural scenes,
our design is tailored for industrial wafer defect images characterized by low contrast, strong periodic textures, and ambiguous local appearances. Accordingly, we replace the original global context aggregation with a lightweight EffectiveSE mechanism and integrate the module into a detection--segmentation framework.
The standard C2f block is replaced by a MUSE module composed of three parallel sub-branches:
a local branch using a $3\times 3$ convolution to capture fine local details,
a surrounding branch using dilated convolution (dilation rate = 2) to model neighborhood context,
An EffectiveSE branch implementing lightweight channel attention via Global Average Pooling (GAP) followed by grouped $1\times 1$ convolution.
Let $\mathbf{x}_{\text{local}}$ and $\mathbf{x}_{\text{dilated}}$ denote the output feature maps of the local and surrounding branches, respectively. The fused contextual feature is computed as:
\begin{equation}
\begin{aligned}
    \mathbf{x}_{\text{ctx}} &= \mathrm{Concat}(\mathbf{x}_{\text{local}}, \mathbf{x}_{\text{dilated}}), \\
    \mathbf{w} &= \sigma\!\left(\mathrm{Conv}_{1\times 1}^{\text{group}}\!\left(\mathrm{GAP}(\mathbf{x}_{\text{ctx}})\right)\right), \\
    \mathbf{F}_B &= \mathbf{w} \odot \mathbf{x}_{\text{ctx}} .
\end{aligned}
\end{equation}
where $\sigma(\cdot)$ denotes the sigmoid function, $\odot$ denotes channel-wise multiplication, and $\mathbf{F}_B$ represents the enhanced feature output of MUSE. This formulation suppresses periodic background interference while amplifying low-contrast defect responses.
\subsection{Multi-Periodic Texture Contrast Enhancement (MPTCE)}
\begin{figure}[t]
    \centering
    \includegraphics[width=1\linewidth]{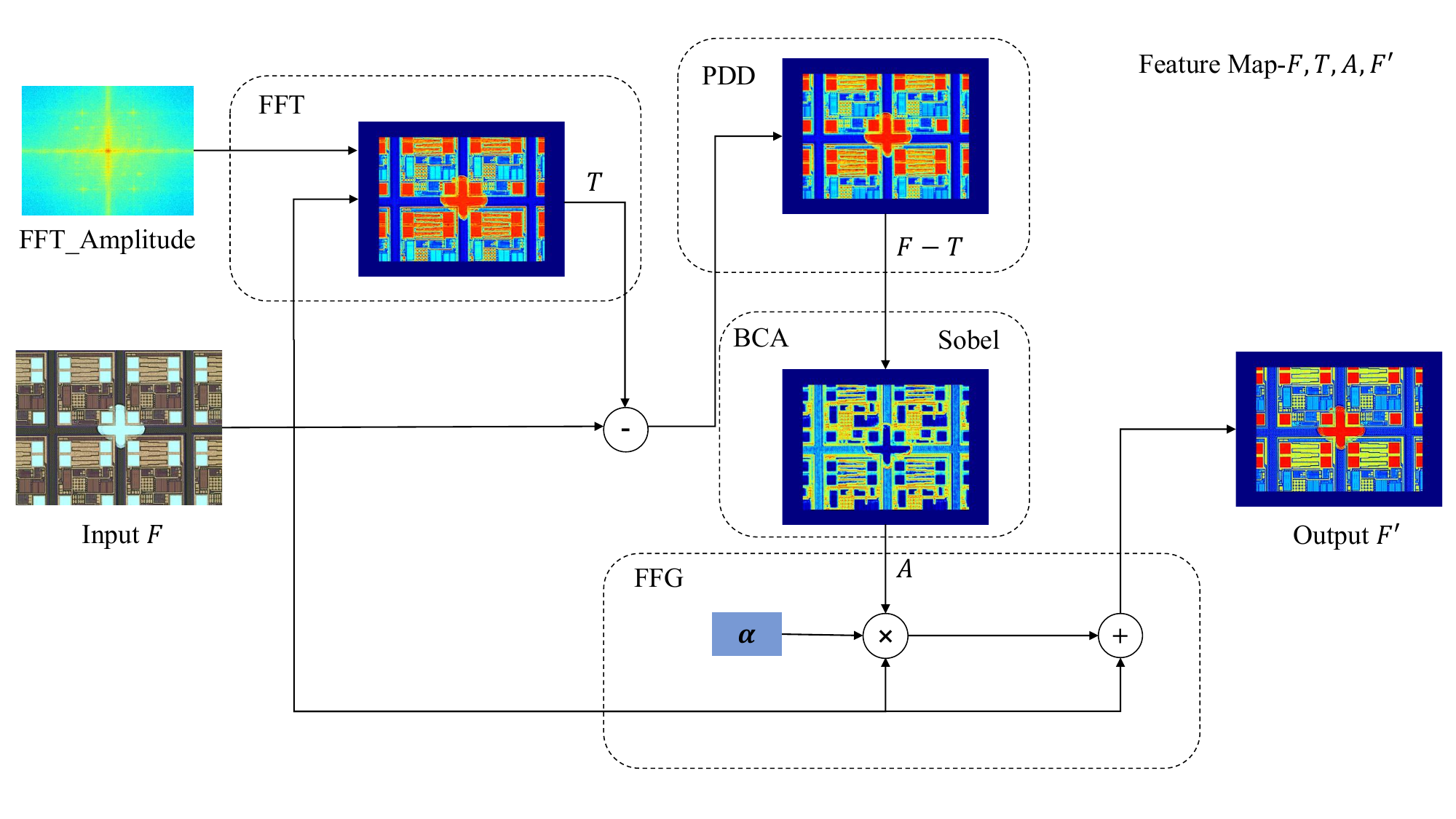}
    \caption{Illustration of frequency decomposition, periodic disturbance extraction, and boundary-consistency attention within the proposed MPTCE module.}
    \label{fig: module_C}
\end{figure}

Wafer surfaces exhibit strong periodic and grating-like structures arising from metal interconnects, repeated cells, and regular grids. Defects typically appear as localized disturbances that disrupt these periodic patterns. Standard spatial convolutions lack an explicit mechanism to model such periodic deviations, motivating the design of a dedicated periodicity-aware enhancement module. As shown in Fig. ~\ref{fig: module_C}, MPTCE operates through three sequential steps.
\noindent\textbf{Periodic Frequency Filter (PFF):}
Given an input feature map $\mathbf{f}(x,y)$ of spatial size $H\times W$, a local frequency transform is applied:
\begin{equation}
    F(u,v) = \sum_{x=1}^{W}\sum_{y=1}^{H} \mathbf{f}(x,y)\,
    e^{-j2\pi\left(\frac{ux}{W}+\frac{vy}{H}\right)} .
\end{equation}
where $(u,v)$ denote frequency indices and $j$ is the imaginary unit. This operation extracts dominant base frequencies corresponding to periodic textures.

\noindent\textbf{Periodic Disturbance Difference (PDD):}
Let $F_{\text{periodic}}$ denote the reconstructed dominant periodic component. The disturbance magnitude is computed as:
\begin{equation}
    \mathbf{D} = \left| F - F_{\text{periodic}} \right| .
\end{equation}
where $\mathbf{D}$ highlights regions where periodicity is disrupted, corresponding to potential defect locations.

\noindent\textbf{Boundary Consistency Attention (BCA):}
Boundary saliency is extracted using a gradient operator:
\begin{equation}
    \mathbf{B} = \nabla \mathbf{D}, \quad
    \mathbf{A} = \sigma\!\left(\mathrm{Conv}(\mathbf{B})\right) .
\end{equation}
where $\mathbf{A}$ denotes the boundary-consistency attention map. The final enhanced feature is obtained as:
\begin{equation}
    \mathbf{F}_C = \mathbf{F} + \alpha\, \mathbf{A} \odot \mathbf{F} .
\end{equation}
where $\mathbf{F}$ denotes the input feature map to MPTCE and $\alpha$ is a learnable scalar controlling the enhancement strength.

\section{Experiment and Analysis}

\subsection{Dataset}
To evaluate the performance of the proposed method, we utilized a self-collected wafer surface defect dataset comprising 4531 RGB images. This dataset encompasses seven typical defect categories commonly encountered in semiconductor manufacturing: block etch, coating bad, particle, polyimide (PI) particle, phosphorus-oxygen (PO) contamination, scratch, and semiconductor edge zone (SEZ) brunt. These defects originate from various manufacturing stages, including etching, coating, contamination deposition, mechanical handling, and thermal reactions, manifesting diverse morphological characteristics on the wafer surface. These images were captured under actual production conditions, thus incorporating variations in lighting, background noise, and wafer structure patterns. The spatial resolution of the dataset ranges approximately from $200\times 200$ to $700\times 700$ pixels, enhancing the robustness of model training. The dataset was randomly split into training, validation, and test subsets at an 8:1:1 ratio. To ensure annotation reliability, all samples were labeled by experienced personnel familiar with wafer inspection standards.

The brief definitions used during annotation are as follows:
block etch: Areas with incomplete or irregular etching due to process non-uniformities.
coating bad: Regions with uneven coating/resist layers, such as streaks, pooling, or pinholes.
particle: Foreign particles visible on the wafer surface (dust, debris).
PI particle: particles attributed to polymer residue after development.
PO contamination: Chemical contaminations exhibiting weak scattering contrast post-oxidation steps.
scratch: Linear mechanical damage marks caused during handling.
SEZ burnt: Local burn or overheating traces associated with the SEZ process.

\subsection{Implementation Details}
To enhance model robustness to varying lighting conditions and structural background noise, several online data augmentation techniques provided by the Ultralytics YOLOv8 training pipeline were employed during training, with all augmentation operations applied solely to the training set. Color jitter (hsv\_h=0.015, hsv\_s=0.70, hsv\_v=0.40) was used to simulate variations in wafer surface reflectivity, along with random translation (translate=0.10) and random scaling (scale=0.50). Horizontal flipping was enabled with a probability of 0.50, while vertical flipping was disabled (flipud=0.0). Four-image mosaic augmentation (mosaic=1.0) was applied in the early training stages (first 10 epochs), and MixUp and Copy-Paste were disabled to avoid creating unrealistic boundary distributions.

All experiments were conducted using the official Ultralytics YOLOv8-seg codebase, with the nano-scale model configuration (yolov8n-seg.yaml) selected as the baseline. The experiments were run on two NVIDIA RTX 4090 GPUs, following a training protocol of 200 epochs with an early-stopping patience of 300, a batch size of 32, and an input resolution of $640\times 640$. The SGD optimizer was used with initial learning rate lr0=0.01, final learning rate factor lrf=0.01, momentum 0.937, and weight decay $5\times 10^{-4}$, along with a 3-epoch warmup period. The loss coefficients were set to box=7.5, cls=0.5, dfl=1.5, and kobj=1.0. For inference, the NMS IoU threshold was 0.7, with a maximum detection number of max\_det=300.

\subsection{Evaluation Metrics}\label{subsec:evaluation_metrics}
To comprehensively evaluate the performance of the proposed TexWDS framework for wafer defect detection and instance segmentation, we adopt a multi-dimensional evaluation protocol aligned with industry standards. The metrics are as follows: Precision, Recall, Mask Intersection over Union (IoU), Average Precision(AP), Mean Average Precision at IoU=0.5 (mAP$_{50}$), Mean Average Precision at IoU=0.5-0.95 (mAP$_{50-95}$), Dice Coefficient, Frames Per Second (FPS), Parameters (Params) and FLOPs. Ensuring a holistic assessment of accuracy, robustness, and deployability.

\begin{equation}
    \text{Precision} = \frac{TP}{TP + FP}.
\end{equation}
\begin{equation}
    \text{Recall} = \frac{TP}{TP + FN}.
\end{equation}
where $TP$, $FP$, and $FN$ denote true positive samples, false positive samples, and false negative samples, respectively. For our task, Precision captures the false alarms arising from low-contrast regions and pseudo-texture backgrounds. In contrast, recall gauges the model’s ability to detect all genuine defects—precision is paramount in industrial defect detection.

\begin{equation}
    \text{IoU} = \frac{ M_{\text{pred}} \cap M_{\text{gt}} }{ M_{\text{pred}} \cup M_{\text{gt}} }.
\end{equation}
where $M_{\text{pred}}$ and $M_{\text{gt}}$ denote the predicted mask and the ground-truth mask, respectively. This metric measures the spatial overlap between the two masks and serves as a direct indicator of segmentation fidelity.

\begin{equation}
\text{AP} = \int_0^1 \text{Precision}(r, \text{IoU}=0.5) \, dr
\end{equation}
\begin{equation}
\text{mAP$_{50}$} = \frac{1}{N} \sum_{c=1}^{N} \text{AP}
\end{equation}
\begin{equation}
    \text{mAP}_{50-95} = \frac{1}{10} \sum_{i=0}^{9} \text{mAP}_{50 + 5i}.
\end{equation}
where $r$ denotes the Recall rate and $N$ denotes the total number of classes, models 7 in this task. They are all used to reflect the localization accuracy of detection boxes and segmentation accuracy.

\begin{equation}
    \text{Dice} = \frac{2 \, |M_{\text{pred}} \cap M_{\text{gt}}| }{ |M_{\text{pred}}| + |M_{\text{gt}}| }.
\end{equation}
where $|\cdot|$ denotes the cardinality of a pixel set. This metric is closely related to IoU but is more sensitive to boundary inaccuracies and partial overlaps, making it especially suitable for evaluating low-contrast contamination or diffuse defects. Its symmetric formulation also enhances interpretability in industrial quality control contexts.

FPS: Frames per second measured on a standardized hardware platform.
Params: Total number of trainable parameters, reflecting model complexity and memory footprint.
FLOPs: Floating-point operations per image, indicating computational cost.

\subsection{Quantitative Results}
\begin{table}[t]
\caption{Comparison of detection and inference metrics between YOLOv8-seg and TexWDS (Ours). (Note: $\uparrow$ indicates higher values are better.)}
\label{tab:detection_comparison}
\centering
\adjustbox{max width=\linewidth,center}{ 
\begin{tabular}{lcccccc}
\toprule
\textbf{Method} & \textbf{Recall $\uparrow$} & \textbf{Precision $\uparrow$} & \textbf{mAP$_{50}$ $\uparrow$} & \textbf{mAP$_{50-95}$ $\uparrow$} & \textbf{Params (M) $\downarrow$} & \textbf{FLOPs (G) $\downarrow$} \\
\midrule
\textbf{YOLOv8-seg (Baseline)} & 0.712 & 0.714 & 0.727 & 0.479 & \textbf{11.2M} & \textbf{28.6} \\
\textbf{TexWDS (Ours)} & \textbf{0.789} & \textbf{0.797} & \textbf{0.804} & \textbf{0.562} & 11.6M & 40.0 \\
\bottomrule
\end{tabular}
}
\end{table}

\begin{table}[t]
\caption{Comparison of segmentation performance between YOLOv8-seg (Baseline) and TexWDS (Ours). (Note: $\uparrow$ indicates higher values are better.)}
\label{tab:segmentation_comparison}
\centering
\adjustbox{max width=\linewidth,center}{
\begin{tabular}{lS[table-format=1.3]S[table-format=1.3]S[table-format=1.3]S[table-format=1.3]S[table-format=1.3]S[table-format=1.3]}
\toprule
\textbf{Method} & \textbf{mask recall$\uparrow$} & \textbf{mask precision$\uparrow$} & \textbf{mask mAP$_{50}$ $\uparrow$} & \textbf{mask mAP$_{50-95}$ $\uparrow$} & \textbf{mask IoU $\uparrow$} & \textbf{Dice $\uparrow$} \\
\midrule
\textbf{YOLOv8-seg (Baseline)} & 0.589 & 0.719 & 0.611 & 0.330 & 0.334 & 0.501 \\
\textbf{TexWDS (Ours)} & \textbf{0.668} & \textbf{0.782} & \textbf{0.753} & \textbf{0.446} & \textbf{0.350} & \textbf{0.518} \\
\bottomrule
\end{tabular}
}
\end{table}

As shown in Tables \ref{tab:detection_comparison} and \ref{tab:segmentation_comparison}, these experiments compare the performance of the TexWDS model and YOLOv8-seg, verifying TexWDS's superiority in two dimensions: object detection and defect segmentation accuracy. The TexWDS model significantly outperforms YOLOv8-seg across four core detection metrics. Meanwhile, the number of parameters increases slightly from 11.2M to 11.6M, and the computational complexity rises from 28.6G to 40.0G. In terms of segmentation performance, TexWDS also achieves superior performance across all metrics compared to YOLOv8-seg.
The aforementioned results demonstrate that TexWDS reduces missed detections and false positives in wafer defect detection tasks. In segmentation tasks, it decreases under-labeling of defect regions and enables more accurate contour delineation. In particular, it exhibits superior comprehensive performance across multiple IoU thresholds and a stronger capability to handle tiny defects with blurred boundaries. Achieving overall performance improvement at the cost of a slight increase in computational complexity.

The P2 branch provides higher-resolution features that compensate for the fine-grained information lost in the P3 layer of YOLOv8-seg, effectively capturing subpixel-level tiny defects and contributing to a 7.7\% improvement in recall.
The MUSE reduces false detections caused by background pseudo-textures through local texture capture, contextual awareness via the dilated convolution branch, and background suppression using channel attention.
The MPTCE explicitly models the periodic background of wafers through mechanisms such as frequency-domain feature extraction and periodic perturbation differences, thereby enhancing the response to low-contrast defects.
The all-round improvement in segmentation performance is mainly attributed to the P2 branch's preservation of fine-grained edge information, which makes defect boundaries sharper and masks more consistently with the actual defect shapes. Additionally, the enhanced ability of the MUSE and MPTCE modules to distinguish defect regions from the background further optimizes the accuracy of segmentation masks.
The slight increase in the number of parameters stems from the branch's sightweight design of the newly added components, which does not lead to significant model expansion and ensures feasibility for engineering deployment. The reasonable rise in computational complexity is a necessary trade-off for performance improvement. In industrial defect-detection scenarios that demand high accuracy, the value of such performance gains far outweighs the impact of increased computational cost. This fully verifies the rationality and superiority of the TexWDS model's structural design.

\begin{table}[t]
\caption{Comparison of per-class mAP$_{50}$ between YOLOv8-seg and TexWDS (Ours) over seven wafer defect categories. (Note: $\uparrow$ indicates higher values are better.)}
\label{tab:per_class_mAP50}
\centering
\adjustbox{max width=\linewidth,center}{ 
\setlength{\tabcolsep}{4pt}
\renewcommand{\arraystretch}{1.2}
\begin{tabular}{lccccccc}
\toprule
\multirow{2}{*}{\textbf{Method}} & \multicolumn{7}{c}{\textbf{mAP$_{50}$}$\uparrow$} \\
\cline{2-8}
& \textbf{block etch} & \textbf{coating bad} & \textbf{particle} & \textbf{PI particle} & \textbf{PO contamination} & \textbf{scratch} & \textbf{SEZ burnt} \\
\midrule
\textbf{YOLOv8-seg (Baseline)} & 0.783 & 0.726 & \textbf{0.829} & 0.669 & 0.761 & 0.812 & 0.742 \\
\textbf{TexWDS (Ours)}     & \textbf{0.813} & \textbf{0.809} & 0.827 & \textbf{0.746} & \textbf{0.823} & \textbf{0.886} & \textbf{0.821} \\
\bottomrule
\end{tabular}
}
\end{table}

\begin{table}[t]
\caption{Comparison of per-class mAP$_{50-95}$ between YOLOv8-seg and TexWDS (Ours) over seven wafer defect categories. (Note: $\uparrow$ indicates higher values are better.)}
\label{tab:per_class_mAP50_95}
\centering
\adjustbox{max width=\linewidth,center}{ 
\setlength{\tabcolsep}{4pt}
\renewcommand{\arraystretch}{1.2}
\begin{tabular}{lccccccc}
\toprule
\multirow{2}{*}{\textbf{Method}} & \multicolumn{7}{c}{\textbf{mAP$_{50-95}$}$\uparrow$} \\
\cline{2-8}
& \textbf{block etch} & \textbf{coating bad} & \textbf{particle} & \textbf{PI particle} & \textbf{PO contamination} & \textbf{scratch} & \textbf{SEZ burnt} \\
\midrule
\textbf{YOLOv8-seg (Baseline)} & 0.544 & 0.366 & 0.484 & 0.405 & 0.429 & 0.595 & 0.477 \\
\textbf{TexWDS (Ours)}     & \textbf{0.601} & \textbf{0.482} & \textbf{0.510} & \textbf{0.419} & \textbf{0.496} & \textbf{0.657} & \textbf{0.553} \\
\bottomrule
\end{tabular}
}
\end{table}

\begin{figure}[t]
    \centering
    \includegraphics[width=1\linewidth]{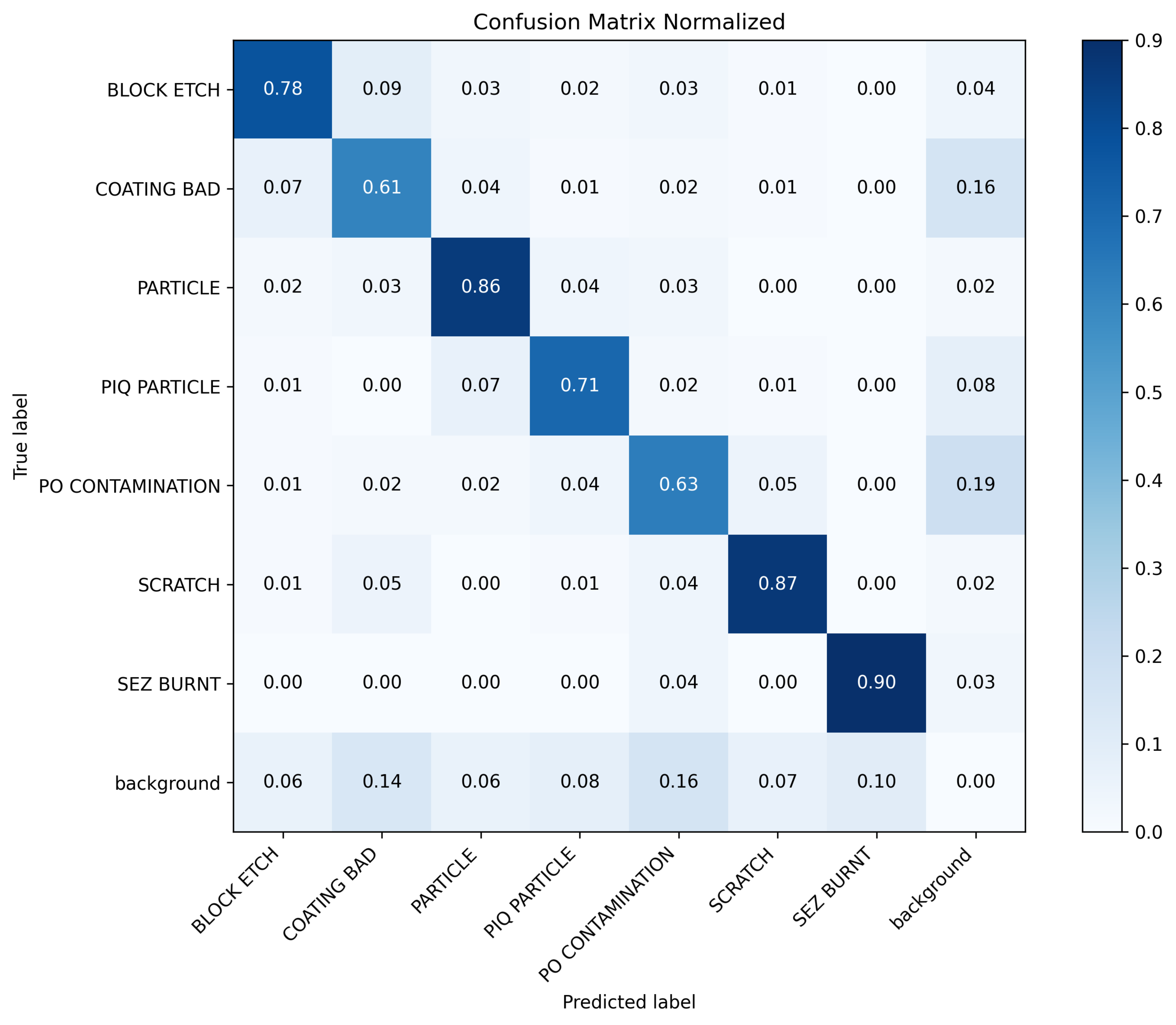}
    \caption{Normalized confusion matrix of the proposed Wafer defect detection framework, TexWDS.}
    \label{fig:confusion_matrix1}
\end{figure}

According to the results presented in Tables \ref{tab:per_class_mAP50}, \ref{tab:per_class_mAP50_95} and Fig.~\ref{fig:confusion_matrix1}. TexWDS achieves performance improvements in six of the seven defect categories, with particularly significant gains observed for PI particle, coating bad, and SEZ burnt defects; the only exception is the particle defect category, where TexWDS performs slightly worse than the baseline. Under the more stringent per-class mAP$_{50-95}$ metric, TexWDS outperforms the baseline across all seven defect categories, with the performance improvements for block etch, coating bad, and SEZ burnt exceeding 10\%.

The aforementioned results demonstrate that TexWDS exhibits stronger detection capability across different types of wafer defects. In particular, it achieves remarkable improvements in detection accuracy for defects with blurred boundaries (block etch, SEZ burnt), medium-sized defects (coating bad, PO contamination), tiny particle defects (PI particle), and scratch defects (scratch). At the same time, its performance remains roughly on par with the baseline only for conventional particle defects. Regarding the causes of these performance discrepancies, for PI particle defects (sub-pixel tiny defects), the substantial improvement in mAP$_{50}$ stems from the P2 microscale branch in TexWDS, which preserves fine-grained edge information, compensates for the loss of tiny target features in the P3 layer of the baseline model, and thus effectively reduces the missed detection rate; the performance gains for medium-sized defects such as coating bad and PO contamination are attributed to the MUSE module's capabilities in local texture capture and background suppression, which reduce false detections caused by background pseudo-textures and enhance the dimodule'sbility of low-contrast regions; the significant improvements in blurred boundary defects (block etch, SEZ burnt) under the mAP$_{50-95}$ metric are primarily driven by the MPTCE module, which explicitly models the periodic background of wafers through mechanisms including frequency-domain feature extraction and periodic perturbation difference, strengthens the texture distinction between defects and regular backgrounds, and thereby improves localization accuracy across multiple IoU thresholds; the notable performance enhancement for scratch defects results from the synergistic effect of the P2 branch (which preserves fine-grained edges of scratches) and the MPTCE module (which suppresses background interference); as for the particle defects, where TexWDS's performance is comparable to the baseline, the reason lies in the fact that such defects exhibit relatively distinct features, and the baseline model aTexWDS'semonstrates satisfactory detection capability, and although TexWDS does not further improve the mAP$_{50}$ for this category, it still achieves an improvement of 0.026 under the more stringent mAP$_{50-95}$ metric, indicating that the model still optimizes the localization accuracy for these defects.

\subsection{Ablation Study}
\begin{table}[t]
\caption{Ablation Study Results on the Contributions of the Three Main Components of the TexWDS Model: P2 Branch, MUSE, and MPTCE. (Note: $\uparrow$ indicates higher values are better.)}
\label{tab:ablation_study}
\centering
\renewcommand{\arraystretch}{1.2}
\setlength{\tabcolsep}{2pt}
\resizebox{\columnwidth}{!}{%
\begin{tabular}{ccc|c|cccc}
\toprule
\textbf{P2 branch} & \textbf{MUSE} & \textbf{MPTCE} & \textbf{Group} & \textbf{Recall $\uparrow$} & \textbf{Precision $\uparrow$} & \textbf{mAP$_{50}$ $\uparrow$} & \textbf{mAP$_{50-95}$ $\uparrow$} \\
\midrule
\ding{55} & \ding{55} & \ding{55} & A0 & 71.2 & 71.4 & 72.7 & 47.9 \\
$\checkmark$ &\ding{55} &\ding{55} & A1 & 74.7 & 73.8 & 74.9 & 50.9 \\
\ding{55}& $\checkmark$ & \ding{55} & A2 & 74.6 & 76.2 & 75.7 & 50.3 \\
\ding{55}& \ding{55}& $\checkmark$ & A3 & 73.7 & 72.4 & 76.2 & 51.1 \\
$\checkmark$ & $\checkmark$ &\ding{55} & A4 & 77.1 & 76.7 & 76.9 & 52.9 \\
$\checkmark$ &\ding{55} & $\checkmark$ & A5 & 75.8 & 77.5 & 76.7 & 51.4 \\
\ding{55}& $\checkmark$ & $\checkmark$ & A6 & 76.6 & 77.1 & 78.7 & 53.5 \\
$\checkmark$ & $\checkmark$ & $\checkmark$ & A7 & \textbf{78.9} & \textbf{79.7} & \textbf{80.4} & \textbf{56.2} \\
\bottomrule
\end{tabular}%
}
\end{table}
The experimental groups (A0–A7) correspond to different combinations of components, where a $\checkmark$ indicates the component is enabled and a blank indicates it is disabled. All experiments are conducted under the same dataset, training hyperparameters, and hardware conditions to ensure a fair comparison. The results are shown in Table \ref{tab:ablation_study}.
All combined models outperform the baseline model (A0), validating the necessity of the core components for enhancing model performance. Specifically, the P2 branch contributes the most to Recall improvement (+3.5), the MUSE module contributes the most to Precision improvement (+4.8), and the MPTCE module contributes the most to mAP$_{50}$ improvement (+3.5). The performance of groups A4-A6 is superior to their corresponding single-component groups, with the MUSE+MPTCE combination (A6) achieving the best results among dual-component combinations, with mAP$_{50}$ of 78.7 and mAP$_{50-95}$ of 53.5. Our proposed TexWDS model (A7) achieves peak values across all metrics, with Recall, Precision, mAP$_{50}$, and mAP$_{50-95}$ reaching 78.9, 79.7, 80.4, and 56.2, respectively, significantly outperforming all other groups.

\textbf{P2 + MUSE (A4)} acts on different visual problems and exhibits complete orthogonality. The P2 branch focuses on the scale problem, addressing micro-defect miss detection to improve Recall; the MUSE module focuses on the contrast problem, addressing background-interference false detection to improve Precision. With no functional crecallts, their combination achieves dual optimization of reduced miss rate and reduced false alarm rate, resulting in precision (77.1) and Precision (76.7) that far exceed those of single-component groups, with comprehensive improvements in mAP metrics.
\textbf{MUSE + MPTCE (A6)} is the optimal dual-component combination in terms of mAP, forming functional complementarity in low-contrast defect detection. The MUSE module provides global contextual perception, helping the model grasp the overall defect distribution; the MPTCE module provides explicit high-frequency information on periodic perturbations, accurately distinguishing defects from periodic backgrounds. After combining both, the response capability to low-contrast defects, such as contamination and blurry boundaries, is significantly enhanced, resulting in an mAP$_{50}$ of 78.7.
The \textbf{Full model (A7)} achieves optimal performance across all metrics because the three components form a complete system of hierarchical optimisation and closed-loop synergy, operating in different feature spaces without functional conflicts. The specific logic is as follows: the P2 branch acts as the foundational layer, solving the problem of whether micro-defects can be detected, laying the foundation for high Recall; the MUSE module acts as the optimization layer, solving the problem of whether the detected object is a defect, ensuring high Precision; threcallE module acts as the enhancement layer, solving the problem of whether defect localization is accurate, improving detection accuprecisionss multiple thresholds. After stacking all three, the model simultaneously possesses the capabilities of full-scale coverage, low false- and miss-detection, and high-precision localization. The results of the ablation experiments fully verify the structural design logic of the TexWDS model, which features "hierarchical optimization and collaborative complementarity", thereby providing solid experimental support for the model's effectiveness and superiority.

\subsection{Visualization Analysis}
\begin{figure}[t]
    \centering
    \includegraphics[width=0.8\linewidth]{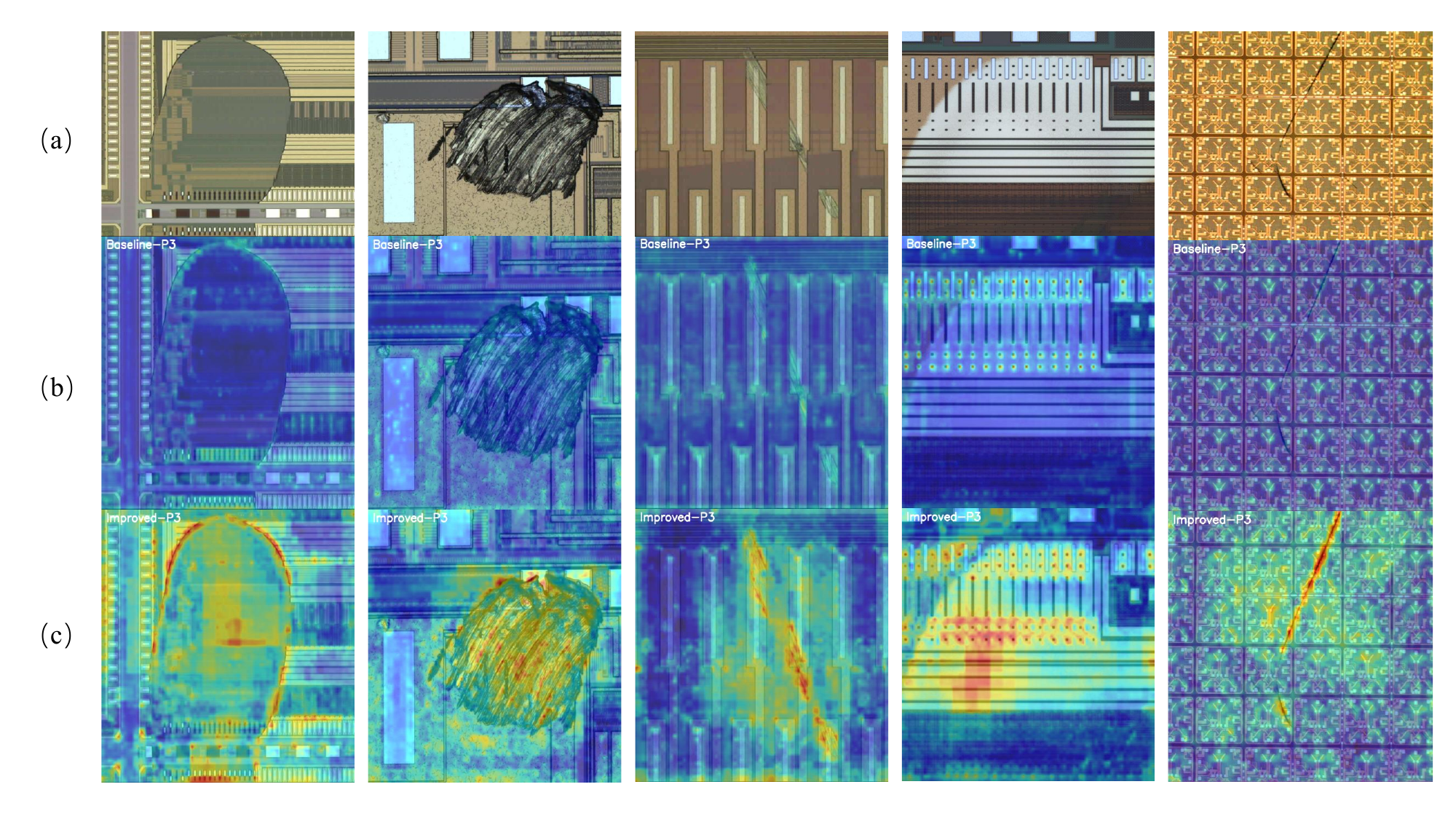}
    \caption{Comparison of feature activation maps in the final P3 detection layer between the baseline model and TexWDS. (a) Original image, (b) YOLOv8, and (c) TexWDS (Ours).}
    \label{fig:visualABC}
\end{figure}
\begin{figure}[t]
    \centering
    \includegraphics[width=0.8\linewidth,trim=10 50pt 10 50pt, clip]{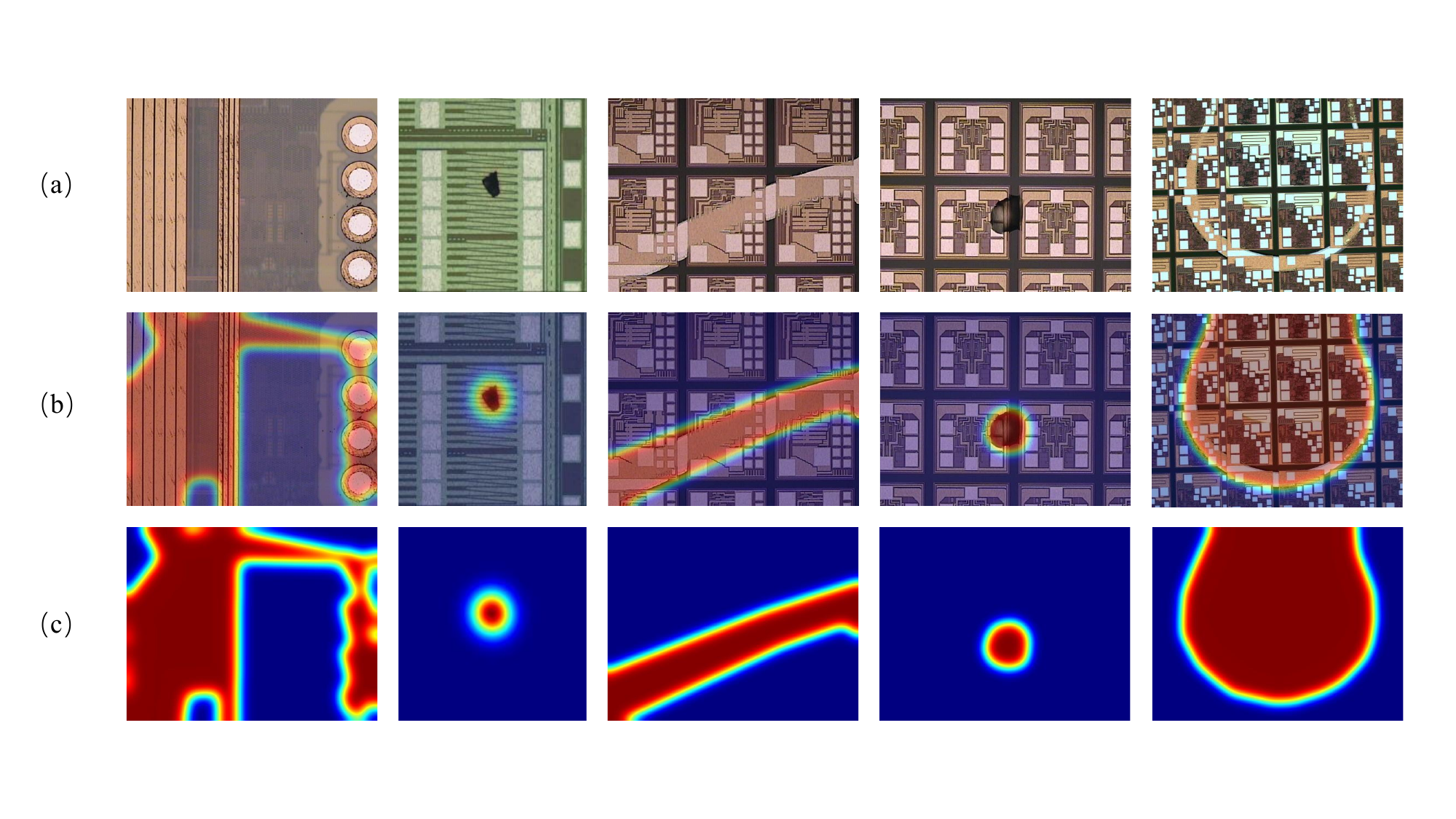}
    \caption{Visualization of periodic disturbance maps generated by the MPTCE module. (a) represents the original image, (b) represents the effect of overlaying the heatmap on the original image, and (c) represents the pure heatmap.}
    \label{fig:MPTCEvisual}
\end{figure}
The comparison of feature activations in the detection layer is shown in Fig.~\ref{fig:visualABC}. Our model exhibits significantly better performance than the baseline. This indicates that the P2 branch enhances fine-grained texture details in shallow layers, the MUSE module suppresses repetitive background patterns, and the MPTCE module amplifies responses at locations where periodicity is disrupted. By integrating these effects, the full model generates the most salient and spatially precise activation over defect regions.

The PDD feature maps are shown in Fig.~\ref{fig:MPTCEvisual}. These heatmaps are directly derived from the difference branch of the MPTCE module and serve as the core intermediate representation for distinguishing between backgrounds and defects. From these visualizations, we infer that the PDD heatmaps exhibit highly structured spatial responses, and their distribution is highly consistent with the physical characteristics of periodic texture disturbances on the wafer surface.
The blue regions primarily correspond to regular, repetitive background textures. This may be attributed to the fact that their frequency-domain features can be effectively modelled by the PFF branch and well-matched to the periodic template. Therefore, after subtraction with the periodic template, these regions show low responses in the PDD feature maps, indicating that they are explicitly identified as periodic backgrounds. The red regions correspond to structural disturbances that cannot be explained by the periodic model, i.e., real defect regions. Since defects brdisrupthe original periodic structure, their frequency-domain features differ significantly from ththose of e template. Such differences are explicitly amplified during the Periodic Disturbance Difference calculation, and ultimately manifest as high response values in the heatmaps.

\begin{figure}[t]
    \centering
    \includegraphics[width=0.8\linewidth,trim=100 10pt 100 10pt, clip]{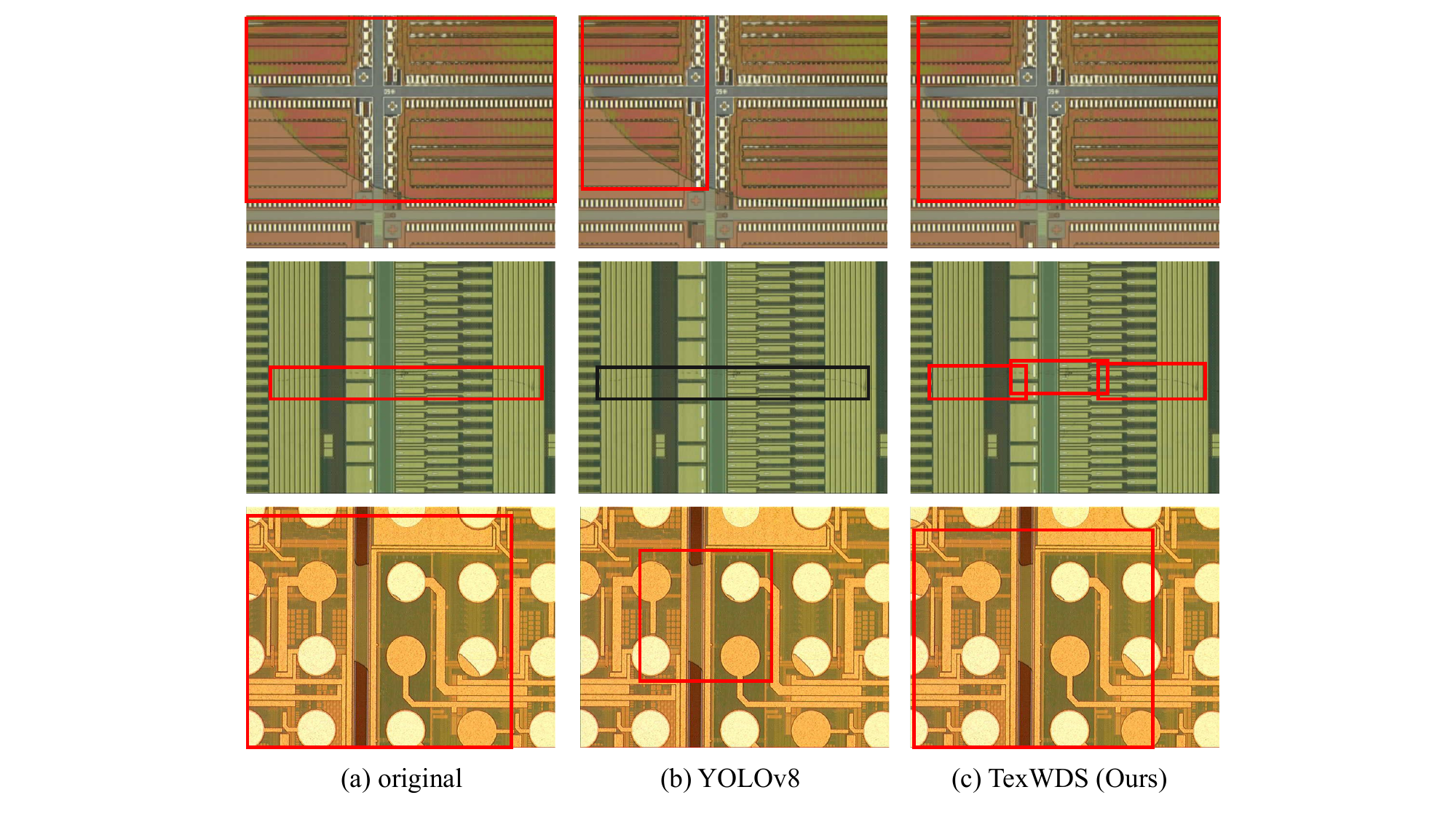}
    \caption{Comparison of detection results. (a), (b) and (c) represent the manually annotated defect bounding boxes and the detection results of YOLOv8 and TexWDS, respectively. Red boxes indicate detected defects, black boxes indicate detection failures.}
    \label{fig:visualbbox}
\end{figure}
\begin{figure}[t]
    \centering
    \includegraphics[width=0.8\linewidth,trim=100 10pt 100 10pt, clip]{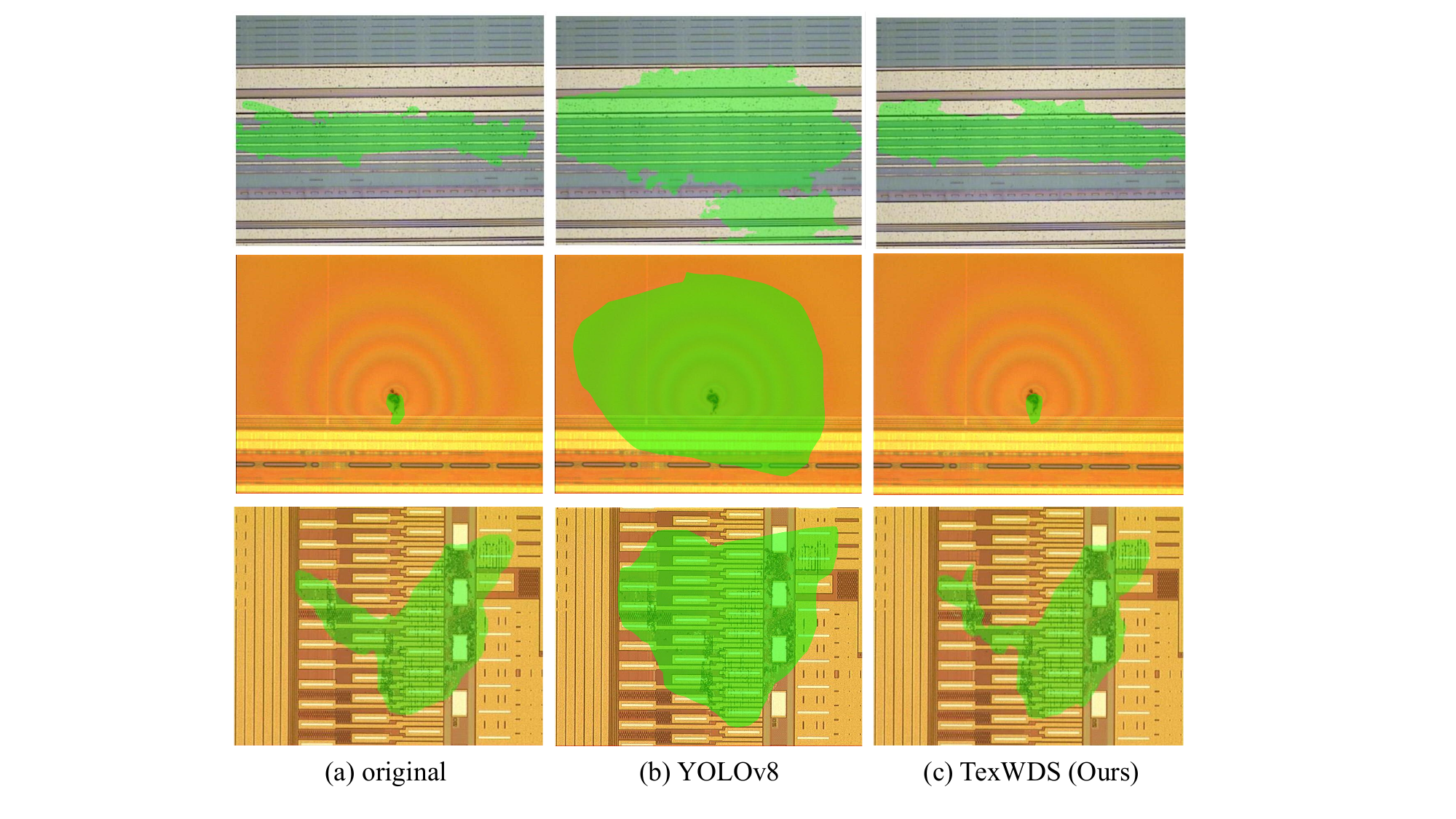}
    \caption{Comparison of Segmentation Results. (a), (b) and (c) denote the manually annotated defect mask, the segmentation result of YOLOv8, and the segmentation result of TexWDS, respectively.}
    \label{fig:visualmask}
\end{figure}
The visualization detection comparison in Fig.~\ref{fig:visualbbox} focuses on industrial hard cases: low-contrast conditions, extremely slender defects, and periodic texture backgrounds.
When grayscale or texture differences between defthe ects and background are weak, the baseline often fails to form stable discriminative releading toses, with significant uncertthe ainty in attention distribution near defect eleads to anThis instability causes insufficient defedetectionsse and missed detection. In contrast, TexWDS maintains a high detection recall rate, with detection boxes more accurately covering real defect regions. This improvement is due to the Context-Guided module’s adaptive channel response modulation, enabling the model to focus on discriminative edge and texture cues even under low contrast. For extremely narrow, linearly distributed scratch defects, the baseline fails to maintain continuous spatial activation in deep feature maps. Such defects easily lose information during downsampling, leading to broken or missing detection boxes. By introducing high-resolution shallow feature enhancement, Tthe exWDS significantly impofes spatial representaenabling slender structures, allowing the model to treat scratches as continuous tar andets rather than random noise, effectively reducing missed detections. In regions with strong periodicground tonly extures, the baseline exhibits clear edge-detectionefect or has obvioa us edge detection loss, indicating failure to distinguish norma,lbroken-periodicityes from abnormal periodicity-broken regions.TexWDS achieves more complete and stable detection results here. Visual analysis shows MPTCE’s BCA explicitly highlights periodicity-broken positions and suppresses repetitive background interference, focusing the model’s attention on real defect boundaries. Thus, the final detection boxes are more complete and accurately aligned with the ground truth.

Through visual comparison in Fig.~\ref{fig:visualmask}, the performance of different models in defect boundary delineation can be clearly observed.
For small defects with slender boundaries, e.g., fine particles or slight scratches, the baseline often fails to maintain stable spatial responses in downsampled feature space. This causes the model to incorporate surrounding high-frequency textures or random noise into predictions, resulting in obvious mask expansion.
In contrast, the P2 microscale featuconstraints onificantly enhances spatial constra,int for micro-structures by using shallow high-resolution features. This makes predicted masks more compact and effectively reduces background false detection.
In regions with gradual boundaries or halo effects, e.g., module's coating defects or slight surface contamination, the baseline frequently misclassifies them as large-area contamination. This problem arises from insufficient discrimination between defects and backgrounds under low contrast.
With MUSE integration, the model can adaptively modulate channel responses using context information. This effectively suppresses false activations caused by slow background brightness changes and significantly reduces such mis-detections.
In wafer regions with strong periodic textures, the baseline tends to misidentify regular texturthat are es as defects, leading to the predicted masks significantly larger than actual areas. This phenomenon is particularly obBy leveragingodic structures are slightly distdetectenefiting from MPTCE, the model can explicitly perceive damaged periodic positions and accurately constrain boundary-consistencyeriodic disturbance difference and boundary consistency attention. Results demonstrate that TexWDS'sE significantly reduces redundant mask regiothe ns here, making predictions highly consistent with actual defect morphology.

\section{Conclusion}
This study addresses three core challenges in wafer defect detection—sub-pixel-scale small objects, low-contrast contamination defects, and interference from periodic textures—by proposing TexWDS, a structured enhancement framework specifically tailored to the intrinsic characteristics of wafer images. The framework achieves multi-dimensional modeling of geometric details, contextual consistency, and periodic texture disruptions through three lightweight components: a P2 microscale feature branch, an MUSE module, and an MPTCE module.
Experimental results demonstrate significant performance gains across seven wafer defect datasets:
The final model improves mAP$_{50-95}$ by 5.8 percentage points and recall by 6.2 percentage points over the original YOLOv8-seg.
Mask IoU increases by 4.9 percentage points, with notably reduced boundary jitter for large-area contamination defects.
For low-contrast deffalse-positive embedded in periodic texture30\% he false positive rate drops by approximately 30\%.

These results validate three key advantages of our framework: strong generalization, high efficiency, and good interpretability. Future work can proceed along two directions: extending the MPTCE component to cross-modal wafer data (e.g., SEM and optical imaging) and exploring weakly supervised or continual learning strategies.

\section*{CRediT authorship contribution statement}
\textbf{Zihan Zhang:} Methodology, Software, Writing – original draft. \textbf{Shijiao Li:} Validation, Investigation. \textbf{Danhang Niu:} Investigation. \textbf{Wei Peng:} Resources. \textbf{Yifan Hu:} Formal analysis. \textbf{Mingfu Zhu:} Data curation. \textbf{Rui Xi:} Funding acquisition. \textbf{Lixin Zhang:} Project administration. \textbf{Tien-Chien Jen:} Conceptualization. \textbf{Ronghan Wei:} Writing – review \& editing, Visualization, Supervision.

\section*{Declaration of Competing Interest}
The authors declare that they have no known competing financial interests or personal relationships that could have appeared to influence the work reported in this paper.

\section*{Acknowledgements}
This work was supported by the National Natural Science Foundation of China (Grant No. 52171193), the Key Research and Development Program of Henan Province (Grant Nos. 241111220400), High-end Foreign Experts Recruitment Program of Henan Province (Grant No. HNGD2023001), and High-end Foreign Experts Recruitment Program of China (G2023026020L).

\section*{Data availability}
The data that support the findings of this study are available from the corresponding authors upon reasonable request.

\bibliography{ref}

@ARTICLE{8263132,
  author={Nakazawa, Takeshi and Kulkarni, Deepak V.},
  journal={IEEE Transactions on Semiconductor Manufacturing}, 
  title={Wafer Map Defect Pattern Classification and Image Retrieval Using Convolutional Neural Network}, 
  year={2018},
  volume={31},
  number={2},
  pages={309-314},
  doi={10.1109/TSM.2018.2795466}}

@Article{mi14050905,
AUTHOR = {Zheng, Jiebing and Zhang, Tao},
TITLE = {Wafer Surface Defect Detection Based on Background Subtraction and Faster R-CNN},
JOURNAL = {Micromachines},
VOLUME = {14},
YEAR = {2023},
NUMBER = {5},
ARTICLE-NUMBER = {905},
URL = {https://www.mdpi.com/2072-666X/14/5/905},
ISSN = {2072-666X},
DOI = {10.3390/mi14050905}
}

@inproceedings{10.1117/12.3033091,
author = {Jianxin Diao and Longchuan Zou and Guihong Zhang},
title = {{Wafer defect detection based on improved YOLOv8}},
volume = {13184},
booktitle = {Third International Conference on Electronic Information Engineering and Data Processing (EIEDP 2024)},
editor = {M. A. Jabbar and Pascal Lorenz},
organization = {International Society for Optics and Photonics},
publisher = {SPIE},
pages = {1318473},
year = {2024},
doi = {10.1117/12.3033091},
URL = {https://doi.org/10.1117/12.3033091}
}

@INPROCEEDINGS{10295689,
  author={Tao, Qian and Chen, Yiyang and Chen, Hongtian},
  booktitle={2023 CAA Symposium on Fault Detection, Supervision and Safety for Technical Processes (SAFEPROCESS)}, 
  title={A Detection Approach for Wafer Detect in Industrial Manufacturing based on YOLOv8}, 
  year={2023},
  volume={},
  number={},
  pages={1-6},
  doi={10.1109/SAFEPROCESS58597.2023.10295689}}

@article{article,
author = {Ma, Jianhong and Zhang, Tao and Yang, Cong and Cao, Yangjie and Xie, Lipeng and Tian, Hui and Li, Xuexiang},
year = {2023},
month = {04},
pages = {1787},
title = {Review of Wafer Surface Defect Detection Methods},
volume = {12},
journal = {Electronics},
doi = {10.3390/electronics12081787}
}

@article{10.1109/TPAMI.1986.4767851,
author = {Canny, J},
title = {A Computational Approach to Edge Detection},
year = {1986},
issue_date = {June 1986},
publisher = {IEEE Computer Society},
address = {USA},
volume = {8},
number = {6},
issn = {0162-8828},
url = {https://doi.org/10.1109/TPAMI.1986.4767851},
doi = {10.1109/TPAMI.1986.4767851},
journal = {IEEE Trans. Pattern Anal. Mach. Intell.},
month = jun,
pages = {679–698},
numpages = {20}
}

@article{haralick1973textural,
  title={Textural Features for Image Classification},
  author={Haralick, Robert M. and Shanmugam, K. and Dinstein, Itshak},
  journal={IEEE Transactions on Systems, Man, and Cybernetics},
  volume={SMC-3},
  number={6},
  pages={610--621},
  year={1973},
  doi={10.1109/TSMC.1973.4309314}
}

@misc{redmon2018yolov3incrementalimprovement,
      title={YOLOv3: An Incremental Improvement}, 
      author={Joseph Redmon and Ali Farhadi},
      year={2018},
      eprint={1804.02767},
      archivePrefix={arXiv},
      primaryClass={cs.CV},
      url={https://arxiv.org/abs/1804.02767}, 
}

@misc{bochkovskiy2020yolov4optimalspeedaccuracy,
      title={YOLOv4: Optimal Speed and Accuracy of Object Detection}, 
      author={Alexey Bochkovskiy and Chien-Yao Wang and Hong-Yuan Mark Liao},
      year={2020},
      eprint={2004.10934},
      archivePrefix={arXiv},
      primaryClass={cs.CV},
      url={https://arxiv.org/abs/2004.10934}, 
}

@misc{lin2017featurepyramidnetworksobject,
      title={Feature Pyramid Networks for Object Detection}, 
      author={Tsung-Yi Lin and Piotr Dollár and Ross Girshick and Kaiming He and Bharath Hariharan and Serge Belongie},
      year={2017},
      eprint={1612.03144},
      archivePrefix={arXiv},
      primaryClass={cs.CV},
      url={https://arxiv.org/abs/1612.03144}, 
}

@misc{chen2017rethinkingatrousconvolutionsemantic,
      title={Rethinking Atrous Convolution for Semantic Image Segmentation}, 
      author={Liang-Chieh Chen and George Papandreou and Florian Schroff and Hartwig Adam},
      year={2017},
      eprint={1706.05587},
      archivePrefix={arXiv},
      primaryClass={cs.CV},
      url={https://arxiv.org/abs/1706.05587}, 
}

@misc{zhang2018imagesuperresolutionusingdeep,
      title={Image Super-Resolution Using Very Deep Residual Channel Attention Networks}, 
      author={Yulun Zhang and Kunpeng Li and Kai Li and Lichen Wang and Bineng Zhong and Yun Fu},
      year={2018},
      eprint={1807.02758},
      archivePrefix={arXiv},
      primaryClass={cs.CV},
      url={https://arxiv.org/abs/1807.02758}, 
}

@misc{cao2020globalcontextnetworks,
      title={Global Context Networks}, 
      author={Yue Cao and Jiarui Xu and Stephen Lin and Fangyun Wei and Han Hu},
      year={2020},
      eprint={2012.13375},
      archivePrefix={arXiv},
      primaryClass={cs.CV},
      url={https://arxiv.org/abs/2012.13375}, 
}

@article{YAN2024e30889,
title = {Convolutional neural network with parallel convolution scale attention module and ResCBAM for breast histology image classification},
journal = {Heliyon},
volume = {10},
number = {10},
pages = {e30889},
year = {2024},
issn = {2405-8440},
doi = {https://doi.org/10.1016/j.heliyon.2024.e30889},
url = {https://www.sciencedirect.com/science/article/pii/S2405844024069202},
author = {Ting Yan and Guohui Chen and Huimin Zhang and Guolan Wang and Zhenpeng Yan and Ying Li and Songrui Xu and Qichao Zhou and Ruyi Shi and Zhi Tian and Bin Wang}
}

@article{INGLE2025103158,
title = {Deep learning driven silicon wafer defect segmentation and classification},
journal = {MethodsX},
volume = {14},
pages = {103158},
year = {2025},
issn = {2215-0161},
doi = {https://doi.org/10.1016/j.mex.2025.103158},
url = {https://www.sciencedirect.com/science/article/pii/S2215016125000068},
author = {Rohan Ingle and Aniket K. Shahade and Mayur Gaikwad and Shruti Patil}
}

@ARTICLE{9292449,
  author={Wu, Tianyi and Tang, Sheng and Zhang, Rui and Cao, Juan and Zhang, Yongdong},
  journal={IEEE Transactions on Image Processing}, 
  title={CGNet: A Light-Weight Context Guided Network for Semantic Segmentation}, 
  year={2021},
  volume={30},
  number={},
  pages={1169-1179},
  doi={10.1109/TIP.2020.3042065}}

@article{Kim2023,
  title={Advances in machine learning and deep learning applications towards wafer map defect recognition and classification: a review},
  author={Kim, Tongwha and Behdinan, Kamran},
  journal={Journal of Intelligent Manufacturing},
  volume={34},
  number={8},
  pages={3215--3247},
  year={2023},
  month={12},
  publisher={Springer},
  issn={1572-8145},
  doi={10.1007/s10845-022-01994-1},
  url={https://doi.org/10.1007/s10845-022-01994-1}
}

@article{shen2025imagedit,
  title={IMAGEdit: Let Any Subject Transform},
  author={Shen, Fei and Xu, Weihao and Yan, Rui and Zhang, Dong and Shu, Xiangbo and Tang, Jinhui},
  journal={arXiv preprint arXiv:2510.01186},
  year={2025}
}

@article{shen2025imagharmony,
  title={IMAGHarmony: Controllable Image Editing with Consistent Object Quantity and Layout},
  author={Shen, Fei and Du, Xiaoyu and Gao, Yutong and Yu, Jian and Cao, Yushe and Lei, Xing and Tang, Jinhui},
  journal={arXiv preprint arXiv:2506.01949},
  year={2025}
}

@article{shen2025imaggarment,
  title={IMAGGarment-1: Fine-Grained Garment Generation for Controllable Fashion Design},
  author={Shen, Fei and Yu, Jian and Wang, Cong and Jiang, Xin and Du, Xiaoyu and Tang, Jinhui},
  journal={arXiv preprint arXiv:2504.13176},
  year={2025}
}

@inproceedings{shenlong,
  title={Long-Term TalkingFace Generation via Motion-Prior Conditional Diffusion Model},
  author={Shen, Fei and Wang, Cong and Gao, Junyao and Guo, Qin and Dang, Jisheng and Tang, Jinhui and Chua, Tat-Seng},
  booktitle={Forty-second International Conference on Machine Learning},
  year={2025},
  pages={32}
}

@article{shen2024imagpose,
  title={Imagpose: A unified conditional framework for pose-guided person generation},
  author={Shen, Fei and Tang, Jinhui},
  journal={Advances in neural information processing systems},
  volume={37},
  pages={6246--6266},
  year={2024}
}

@inproceedings{shen2025imagdressing,
  title={Imagdressing-v1: Customizable virtual dressing},
  author={Shen, Fei and Jiang, Xin and He, Xin and Ye, Hu and Wang, Cong and Du, Xiaoyu and Li, Zechao and Tang, Jinhui},
  booktitle={Proceedings of the AAAI Conference on Artificial Intelligence},
  volume={39},
  pages={6795--6804},
  year={2025}
}

@article{Terven_2023,
   title={A Comprehensive Review of YOLO Architectures in Computer Vision: From YOLOv1 to YOLOv8 and YOLO-NAS},
   volume={5},
   ISSN={2504-4990},
   url={http://dx.doi.org/10.3390/make5040083},
   DOI={10.3390/make5040083},
   number={4},
   journal={Machine Learning and Knowledge Extraction},
   publisher={MDPI AG},
   author={Terven, Juan and Córdova-Esparza, Diana-Margarita and Romero-González, Julio-Alejandro},
   year={2023},
   month=nov, pages={1680–1716} }

@INPROCEEDINGS{7780444,
  author={Liang, Xiaodan and Wei, Yunchao and Shen, Xiaohui and Jie, Zequn and Feng, Jiashi and Lin, Liang and Yan, Shuicheng},
  booktitle={2016 IEEE Conference on Computer Vision and Pattern Recognition (CVPR)}, 
  title={Reversible Recursive Instance-Level Object Segmentation}, 
  year={2016},
  volume={},
  number={},
  pages={633-641},
  doi={10.1109/CVPR.2016.75}}

@InProceedings{Long_2015_CVPR,
author = {Long, Jonathan and Shelhamer, Evan and Darrell, Trevor},
title = {Fully Convolutional Networks for Semantic Segmentation},
booktitle = {Proceedings of the IEEE Conference on Computer Vision and Pattern Recognition (CVPR)},
month = {June},
year = {2015},
pages = {3431-3440}
}

@InProceedings{Lin_2017_CVPR,
author = {Lin, Tsung-Yi and Dollar, Piotr and Girshick, Ross and He, Kaiming and Hariharan, Bharath and Belongie, Serge},
title = {Feature Pyramid Networks for Object Detection},
booktitle = {Proceedings of the IEEE Conference on Computer Vision and Pattern Recognition (CVPR)},
month = {July},
year = {2017},
pages = {2117--2125}
}

@INPROCEEDINGS{8237467,
  author={Dong, Qi and Gong, Shaogang and Zhu, Xiatian},
  booktitle={2017 IEEE International Conference on Computer Vision (ICCV)}, 
  title={Class Rectification Hard Mining for Imbalanced Deep Learning}, 
  year={2017},
  volume={},
  number={},
  pages={1869-1878},
  doi={10.1109/ICCV.2017.205}}

@inproceedings{shen2024advancing,
title={Advancing Pose-Guided Image Synthesis with Progressive Conditional Diffusion Models},
author={Fei Shen and Hu Ye and Jun Zhang and Cong Wang and Xiao Han and Yang Wei},
booktitle={The Twelfth International Conference on Learning Representations},
year={2024},
pages={14},
url={https://openreview.net/forum?id=rHzapPnCgT}
}

@INPROCEEDINGS{9090668,
  author={Chan, Carmen and Hwang, Angel and Sun, Daphne and Birckhead, Brandon and Won, Andrea Stevenson},
  booktitle={2020 IEEE Conference on Virtual Reality and 3D User Interfaces Abstracts and Workshops (VRW)}, 
  title={Minimal Embodiment: Effects of a Portable Version of a Virtual Disembodiment Experience on Fear of Death}, 
  year={2020},
  volume={},
  number={},
  pages={746-747},
  doi={10.1109/VRW50115.2020.00224}}

@INPROCEEDINGS{7515504,
  author={Maletti, Andreas and Osterholzer, Johannes},
  booktitle={2016 IEEE 46th International Symposium on Multiple-Valued Logic (ISMVL)}, 
  title={On the Complexity of the Equivalence Problem for Deterministic Tree-Walking Automata}, 
  year={2016},
  volume={},
  number={},
  pages={C4-C4},
  doi={10.1109/ISMVL.2016.60}}

@INPROCEEDINGS{8099490,
  author={Chen, Xinlei and Girshick, Ross and He, Kaiming and Doll{\'a}r, Piotr},
  booktitle={2017 IEEE Conference on Computer Vision and Pattern Recognition (CVPR)}, 
  title={Learning to Segment Every Thing}, 
  year={2017},
  volume={},
  number={},
  pages={lvii-lvii},
  doi={10.1109/CVPR.2017.7}}

@INPROCEEDINGS{8578843,
  author={Hu, Jie and Shen, Li and Sun, Gang},
  booktitle={2018 IEEE/CVF Conference on Computer Vision and Pattern Recognition}, 
  title={Squeeze-and-Excitation Networks}, 
  year={2018},
  volume={},
  number={},
  pages={7132-7141},
  doi={10.1109/CVPR.2018.00745}}

@INPROCEEDINGS{8746000,
  author={Jin, Xianing and Zhang, Jian and Qu, Yankun and Song, Yi and Yang, Weihong and Xu, Jing},
  booktitle={2018 IEEE 2nd International Electrical and Energy Conference (CIEEC)}, 
  title={An Evaluation Method of Diversified Load Absorptive Ability in Distribution Network Based on Hierarchical Component Theory}, 
  year={2018},
  volume={},
  number={},
  pages={120-124},
  doi={10.1109/CIEEC.2018.8746000}}

@inproceedings{chen2018encoder,
  title={Encoder-decoder with atrous separable convolution for semantic image segmentation},
  author={Chen, Liang-Chieh and Zhu, Yukun and Papandreou, George and Schroff, Florian and Adam, Hartwig},
  booktitle={Proceedings of the European conference on computer vision (ECCV)},
  pages={801--818},
  year={2018}
}

@article{Ma2023Review,
  title={Review of Wafer Surface Defect Detection Methods},
  author={Ma, Jianhong and Zhang, Tao and Yang, Cong and Cao, Yangjie and Xie, Lipeng and Tian, Hui and Li, Xuexiang},
  journal={Electronics},
  volume={12},
  number={8},
  pages={1787},
  year={2023},
  publisher={MDPI},
  doi={10.3390/electronics12081787}
}

@article{Tyrone2023Benchmarking,
  title={Benchmarking Classical and Deep Learning Approaches for Defect Detection in High-Resolution Wafer Inspection Imaging},
  author={Tyrone, Reginald},
  journal={International Journal of Computer Science (IJCS)},
  volume={3},
  number={1},
  pages={7--12},
  year={2023}
}

@article{LopezDeLaRosa2023SqueezeNet,
  title={Defect detection and classification on semiconductor wafers using geometric data augmentation and SqueezeNet CNN},
  author={L{\'o}pez de la Rosa, Francisco and G{\'o}mez-Sirvent, Jos{\'e} L. and Morales, Rafael and others},
  journal={Computers \& Industrial Engineering},
  volume={183},
  pages={109549},
  year={2023},
  doi={10.1016/j.cie.2023.109549}
}

@article{Dey2022MaskRCNNSEM,
  title={Deep Learning based Defect classification and detection in SEM images: A Mask R-CNN approach},
  author={Bappaditya Dey and Enrique Dehaerne and others},
  journal={arXiv preprint},
  year={2022}
}

@article{LI2023102470RCIM,
title = {Deep learning based online metallic surface defect detection method for wire and arc additive manufacturing},
journal = {Robotics and Computer-Integrated Manufacturing},
volume = {80},
pages = {102470},
year = {2023},
issn = {0736-5845},
doi = {https://doi.org/10.1016/j.rcim.2022.102470},
url = {https://www.sciencedirect.com/science/article/pii/S0736584522001521},
author = {Wenhao Li and Haiou Zhang and Guilan Wang and Gang Xiong and Meihua Zhao and Guokuan Li and Runsheng Li}
}

@article{WANG2023102513RCIM,
title = {Knowledge augmented broad learning system for computer vision based mixed-type defect detection in semiconductor manufacturing},
journal = {Robotics and Computer-Integrated Manufacturing},
volume = {81},
pages = {102513},
year = {2023},
issn = {0736-5845},
doi = {https://doi.org/10.1016/j.rcim.2022.102513},
url = {https://www.sciencedirect.com/science/article/pii/S0736584522001958},
author = {Junliang Wang and Pengjie Gao and Jie Zhang and Chao Lu and Bo Shen}
}

@article{WANG2024102791RCIM,
title = {Model-enabled robotic machining framework for repairing paint film defects},
journal = {Robotics and Computer-Integrated Manufacturing},
volume = {89},
pages = {102791},
year = {2024},
issn = {0736-5845},
doi = {https://doi.org/10.1016/j.rcim.2024.102791},
url = {https://www.sciencedirect.com/science/article/pii/S0736584524000784},
author = {Shengzhe Wang and Ziyan Xu and Yidan Wang and Ziyao Tan and Dahu Zhu}
}

\end{document}